%% file: main.tex
\documentclass[10pt,twocolumn,letterpaper]{article}

\usepackage{cvpr}              

\input{preamble}

\definecolor{cvprblue}{rgb}{0.21,0.49,0.74}
\usepackage[pagebackref,breaklinks,colorlinks,citecolor=cvprblue]{hyperref}
\usepackage{wrapfig}
\usepackage{amsmath}

\def\paperID{5757} 
\def\confName{CVPR}
\def\confYear{2024}

\title{INFAMOUS-NeRF: ImproviNg FAce MOdeling Using Semantically-Aligned Hypernetworks with Neural Radiance Fields}

 \author{First Author\\
 Institution1\\
 Institution1 address\\
 {\tt\small firstauthor@i1.org}
 \and
 Second Author\\
 Institution2\\
 First line of institution2 address\\
 {\tt\small secondauthor@i2.org}
 }

\author{Andrew Hou\thanks{All of the datasets mentioned in this paper were solely downloaded and used by Michigan State University. } $^{,1}$, Feng Liu$^{1}$, Zhiyuan Ren$^{1}$, Michel Sarkis$^{2}$, Ning Bi$^{2}$, Yiying Tong$^{1}$, Xiaoming Liu$^{1}$ \\
{ $^{1}$Michigan State University, $^{2}$Qualcomm Technologies Inc.
} \\
{\tt\small \{houandr1, liufeng6, renzhiy1, ytong, liuxm\}@msu.edu, \{msarkis, nbi\}@qti.qualcomm.com} \\
}

\newcommand{\E}{\mathbb{E}}

\newcommand{\Eb}[2]{\E_{#1}\!\left[#2\right]}

\newcommand{\bx}{\mathbf{x}}

\newcommand{\bepsilon}{{\boldsymbol{\epsilon}}}

\usepackage{amsmath}
\DeclareMathOperator*{\argmax}{arg\,max}

\begin{document}
\maketitle
\input{sec/0_abstract}    
\input{sec/1_intro}

\input{sec/2_related}
\input{sec/3_method}
\input{sec/4_experiments}
\input{sec/5_conclusion}

\input{sec/X_suppl}

{
    \small
    \bibliographystyle{ieeenat_fullname}
    \bibliography{main}
}

\end{document}

%% file: preamble.tex
\usepackage[dvipsnames]{xcolor}


%% file: sec/0_abstract.tex
\begin{abstract}
 We propose INFAMOUS-NeRF, an implicit morphable face model that introduces hypernetworks to NeRF to improve the representation power in the presence of many training subjects. At the same time, INFAMOUS-NeRF resolves the classic hypernetwork tradeoff of representation power and editability by learning semantically-aligned latent spaces despite the subject-specific models, all without requiring a large pretrained model. INFAMOUS-NeRF further introduces a novel constraint to improve NeRF rendering along the face boundary. Our constraint can leverage photometric surface rendering and multi-view supervision to guide surface color prediction and improve rendering near the surface. Finally, we introduce a novel, loss-guided adaptive sampling method for more effective  NeRF training by reducing the sampling redundancy. We show quantitatively and qualitatively that our method achieves higher representation power than prior face modeling methods in both controlled and in-the-wild settings. Code and 
 models will be released upon publication. 
\end{abstract}

%% file: sec/1_intro.tex
\section{Introduction}
$3$D face modeling is a fundamental problem that has received great interest from the computer vision and graphics communities. It has a wide range of applications including volumetric avatars for AR/VR~\cite{Jourabloo_2022_CVPR,park2021hypernerf,park2021nerfies,Gao-portraitnerf,NeRFace, raj2021pva,NHA}, face alignment~\cite{joint-face-alignment-and-3d-face-reconstruction-with-application-to-face-recognition,on-learning-3d-face-morphable-model-from-in-the-wild-images}, and face recognition~\cite{joint-face-alignment-and-3d-face-reconstruction-with-application-to-face-recognition,disentangling-features-in-3d-face-shapes-for-joint-face-reconstruction-and-recognition}. 

Traditional methods~\cite{on-learning-3d-face-morphable-model-from-in-the-wild-images, DFNRMVS,shang2020self,UncertaintyFaceReconstruction,wang2022faceverse} use explicit representations such as meshes coupled with $3$D morphable models ($3$DMMs)~\cite{DECA,FLAME,yenamandra2020i3dmm,3DMM} to enable learning dense $3$D shapes with correspondences. Despite their applicability to downstream tasks, these methods often rely on principal component analysis (PCA) to model shape and texture from limited $3$D scans and suffer from limited representation power. The mesh data structure itself is another limitation, as it is challenging to model features such as hair and eyes in fine detail and handle topological variations. 

\begin{figure}[t]
\vspace{-2mm}
\begin{center}
   \includegraphics[width=\linewidth]{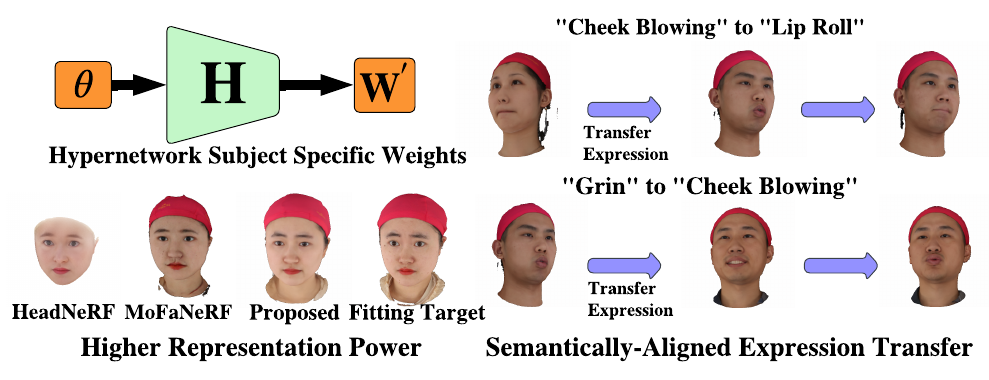}
\vspace{-8mm}
\caption{\small \textbf{Overview}. We propose INFAMOUS-NeRF: a novel face modeling method that improves representation power over prior work and enables various face editing applications. Despite learning subject-specific models through hypernetworks to improve fidelity, INFAMOUS-NeRF learns to realign its latent spaces and can thus perform semantically meaningful face editing across subjects such as expression transfer.   
}\label{fig:Overview}
\end{center}
\vspace{-8mm}
\end{figure}

The introduction of Neural Radiance Fields (NeRF)~\cite{NeRF} substantially improves the rendering quality of face modeling methods. However, due to the complexity of encoding $3$D scenes, NeRF traditionally performs best in the single subject setting~\cite{park2021hypernerf,NeRFace,park2021nerfies,RigNeRF} where it can overfit to data with limited variation. When many subjects are introduced, the representation power of NeRF suffers from the burden of the MLP to memorize many different appearances in the same model. To alleviate this problem, MoFaNeRF~\cite{zhuang2022mofanerf} proposes an implicit $3$DMM that involves a texture encoding module (TEM) conditioned on a UV texture to improve shape and appearance disentanglement, thus allowing the appearance MLP to learn only appearance and not shape-dependent information. 
HeadNeRF initializes its latent codes using an existing 3DMM~\cite{LuanCVPR18} and constrains the optimized latent codes to be close to the initialization, which maintains disentanglement and improves representation power. 
$3$DMM-RF~\cite{3DMM-RF} trains a parametric face model on a synthetic dataset and improves the representation power using SPADE layers~\cite{SPADE}. 
While these methods improve representation power, they do not resolve the fundamental burden of the NeRF MLP in modeling many subjects with one set of weights. 

To resolve this issue, INFAMOUS-NeRF leverages hypernetworks~\cite{hypernetworks} to learn subject-specific NeRF MLP weights instead of a single set of weights, which greatly enhances its representation power. However, hypernetworks carry with them a substantial tradeoff: in return for higher representation power, performing face editing becomes more challenging due to each subject-specific MLP carrying different weights and thus latent spaces with different semantic meanings. We thus propose a method to resolve this problem by enforcing that pairs of images with common semantic attributes (\textit{e.g.}, expressions) learn the same latent code for that attribute. For example, two smiling subjects have the same expression code, which realigns the expression latent space for interpretable face editing across all subjects. Notably, unlike prior hypernetwork methods for face editing~\cite{hyperstyle,hyperinverter} that rely on a pretrained StyleGAN model to maintain editability, INFAMOUS-NeRF can achieve this semantic realignment by training from scratch without the use of a pretrained model, which is important in the NeRF setting as pretrained NeRF MLPs that generalize across many scenes are largely unavailable.     

By observing rendering errors are concentrated at face boundaries, we further propose a novel photometric surface constraint to improve NeRF rendering along the boundary. We achieve this by first identifying the surface point along each ray and ensuring the predicted color at the surface point is consistent with multi-view supervision, thus improving the rendering at pixels near the surface. 

Finally, we propose an adaptive sampling algorithm to improve NeRF rendering quality. Uniform sampling, commonly employed in NeRF, tends to oversample redundant pixels and likewise undersample pixels in regions with challenging textures. This ultimately reduces the rendering quality since more challenging regions are sampled less. To combat this, we propose a novel adaptive sampling algorithm that uses the per-pixel rendering loss to guide where the model should sample in subsequent iterations, with higher loss leading to a higher probability of being sampled.

We demonstrate experimentally that INFAMOUS-NeRF achieves state-of-the-art performance on single-image novel view synthesis and $3$DMM fitting on FaceScape~\cite{yang2020facescape} as well as two in-the-wild datasets FFHQ~\cite{FFHQ} and CelebAHQ~\cite{CelebA-HQ}, outperforming existing face modeling work in representation power and rendering quality. To demonstrate that we are able to learn semantically-aligned latent spaces, we conduct expression transfer experiments that show our learned expression latent space has the same meaning across different subjects despite using hypernetworks. We further demonstrate that our novel photometric surface constraint and adaptive sampling also improve INFAMOUS-NeRF's rendering quality.   

Our contributions are thus fourfold: 

$\diamond$ A novel NeRF-based face modeling method that resolves the tradeoff between representation power and editability when using hypernetworks by learning semantically-aligned latent spaces without the need for a large pretrained model. 

$\diamond$ A novel photometric surface  constraint that improves the rendering quality along the face boundary. 

$\diamond$ A novel ray sampling algorithm for training NeRF that improves rendering quality. 

$\diamond$ State-of-the-art novel view synthesis and fitting performance on multiple benchmarks compared to prior face modeling methods quantitatively and qualitatively.
\label{sec:intro}

%% file: sec/2_related.tex
\section{Related Work}
\label{sec:relatedwork}

\paragraph{Face Modeling.}
Existing face modeling methods can be divided into two groups: explicit methods that rely on mesh representations and PCA bases to model shape and texture~\cite{diffusionrig,LuanCVPR18,towards-high-fidelity-nonlinear-3d-face-morphable-model,TowardsHighFidelityFaceReconstruction,on-learning-3d-face-morphable-model-from-in-the-wild-images, DFNRMVS, riggable-3d-face-reconstruction-via-in-network-optimization, 3d-face-modeling-from-diverse-raw-scan-data,Jiang2019Disentangled,DECA,FLAME,shang2020self,UncertaintyFaceReconstruction,MonocularNeuralReflectance,AU3DRecon,wang2022faceverse,joint-face-alignment-and-3d-face-reconstruction-with-application-to-face-recognition} and implicit methods that rely on NeRF or SDF to model subject appearance and geometry at  high resolutions using per-pixel ray tracing~\cite{zhuang2022mofanerf, yenamandra2020i3dmm, zhang2022fdnerf,hong2021headnerf, NeRFace,CompositionalRF,face-relighting-with-geometrically-consistent-shadows,Gao-portraitnerf, park2021hypernerf, park2021nerfies,raj2021pva,NHA,RigNeRF,EG3D,chanmonteiro2020pi-GAN,GeoD,deng2022gram,GRAF,stylenerf,GIRAFFE,shi2023pof3d,nerfinvertor,Quantmanip,RODIN,PanoHead}. 

While explicit methods have advantages such as dense correspondences, these methods often suffer from a lack of representation power due to their reliance on PCA bases learned from limited $3$D scans. Compared to implicit face modeling methods, especially those that use NeRF as the $3$D representation, the gap in rendering quality is noticeably large. However, despite the higher rendering quality of implicit face modeling methods, they also have limitations with regard to representation power. In the case of NeRF, it is difficult for the MLP to memorize the appearances of many training subjects and the rendering quality noticeably degrades compared to the single subject setting. INFAMOUS-NeRF improves the representation power by introducing hypernetworks to estimate the NeRF MLP, resulting in subject-specific models rather than a single shared model. 
\vspace{-3mm}
\paragraph{Neural Radiance Fields.} As NeRF~\cite{NeRF} involves encoding a high-fidelity photorealistic scene representation, learning a universal NeRF model that maintains high representation power for multiple scenes simultaneously has been a longstanding challenge. Existing work conditions on input images~\cite{yu2020pixelnerf} or partial scenes~\cite{XNeRF} but trains on a limited number of scenes with noticeably worse testing performance compared to single-scene NeRF methods. To alleviate this problem, INFAMOUS-NeRF utilizes hypernetworks~\cite{hypernetworks} to allow for subject-specific NeRF MLPs instead of optimizing a single NeRF MLP for all training subjects.  

\vspace{-3mm}
\paragraph{Hypernetworks.} Hypernetworks have improved the representation power in various domains~\cite{hyperstyle,hyperinverter,hyperseg,deepmeta,Hyperpocket,GraphHN,ContinualLearningHN} by predicting input-specific model weights. However, they also possess an inherent limitation in the image editing domain. As the learned model weights are subject-specific, the latent spaces of the subject-specific models become misaligned, which makes editing challenging. Several methods have tried to address this tradeoff between representation power and editing by leveraging existing well-learned latent spaces. HyperStyle~\cite{hyperstyle} refines the weights of a pretrained StyleGAN generator by using a hypernetwork to predict the layer offsets for each image. HyperInverter~\cite{hyperinverter} uses an encoder to map the input image to the StyleGAN latent space and then uses hypernetworks to refine the StyleGAN generator weights to be compatible with the estimated latent code. While these methods can maintain the original semantic alignment of the StyleGAN latent space, they require a pretrained model with a well-learned latent space. Our method is able to learn semantically-aligned latent spaces from scratch by encouraging similar latent codes for shared attributes between subjects, such as similar expressions. Notably, we do not require a pretrained face modeling method with a large and representative training set, which is usually unavailable for the NeRF architecture due to the difficulty of encoding many scenes simultaneously~\cite{XNeRF,zhuang2022mofanerf}. 

%% file: sec/3_method.tex
\section{Methodology}

\begin{figure*}[t]
\vspace{-2mm}
\begin{center}
   \includegraphics[width=\linewidth]{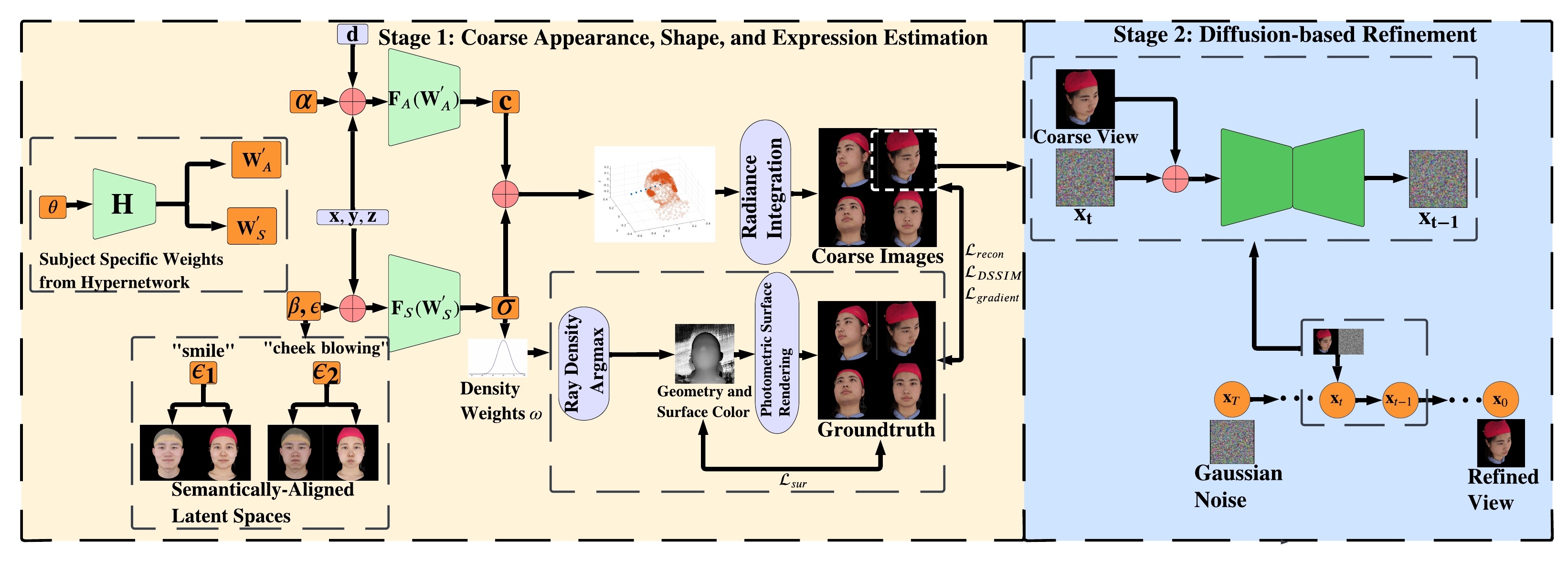}
\vspace{-7mm}
\caption{\small \textbf{Model Architecture}. INFAMOUS-NeRF is a two-stage, single-image face modeling method, where stage $1$ is a NeRF-based morphable model that can coarsely model facial appearance, shape, and expression and stage $2$ is a conditional DDPM refinement. INFAMOUS-NeRF enhances its representation power by estimating subject-specific appearance and shape MLP weights ($\textbf{W}^{'}_{A}$ and $\textbf{W}^{'}_{S}$) using hypernetworks (denoted as $\textbf{H}$) while maintaining editability by learning semantically-aligned latent spaces. INFAMOUS-NeRF further proposes a photometric surface rendering constraint that improves the rendering by focusing on the head boundary. 
}\label{fig:Architecture}
\vspace{-5mm}
\end{center}
\end{figure*}

\subsection{Problem Formulation}
The traditional $3$D face modeling formulation involves mapping multiple latent codes for appearance ($\mathbf{\alpha}$), shape ($\mathbf{\beta}$), and expression ($\mathbf{\epsilon}$) to face texture and geometry using some network $F$. In our case, we have both a shape MLP $F_{S}(\mathbf{W}_{S})$ and an appearance MLP $F_{A}(\mathbf{W}_{A})$ similar to MoFaNeRF~\cite{zhuang2022mofanerf} that model the volume density $\sigma$ and the RGB texture $\mathbf{c}$ respectively at each $3$D location. Following the convention of NeRF~\cite{NeRF}, given a single point $\mathbf{x}$ and a viewing direction $\mathbf{d}$: 
\begin{equation}
    \mathbf{c}=F_{A}(\mathbf{\alpha},\mathbf{x},\mathbf{d}; \mathbf{W}_{A}), \sigma=F_{S}(\mathbf{\alpha}, \mathbf{\beta}, \mathbf{\epsilon}, \mathbf{x}, \mathbf{d}; \mathbf{W}_{S}).
\end{equation}
To improve representation power, instead of learning a universal set of weights $\mathbf{W}_{A}$ and $\mathbf{W}_{S}$ for all subjects, we estimate subject-specific MLP weights $\mathbf{W}^{'}_{A}, \mathbf{W}^{'}_{S}$ by employing hypernetworks, such that: 
\begin{equation}
    \left(\mathbf{W}^{'}_{A}, \mathbf{W}^{'}_{S}\right) = H(\mathbf{\theta}),
\end{equation}
where $H$ is our hypernetwork and $\mathbf{\theta}$ is a learned subject-specific identity latent code. 

\subsection{Proposed Method}
INFAMOUS-NeRF is a two-stage method, where stage $1$ is a NeRF model that estimates coarse appearance, shape, and expression and stage $2$ is a coarse image conditioned DDPM~\cite{DDPM} refinement. We largely adopt the NeRF MLP architecture of MoFaNeRF~\cite{zhuang2022mofanerf} during stage $1$, including $3$ learned latent features $\mathbf{\alpha}$ (appearance), $\mathbf{\beta}$ (shape), and $\mathbf{\epsilon}$ (expression). Unlike MoFaNeRF, we do not learn a single appearance and shape MLP for all training subjects, and instead employ hypernetwork $H$ conditioned on subject-specific identity latent code $\mathbf{\theta}$ to estimate subject-specific appearance and shape MLP weights $\mathbf{W}^{'}_{A}$ and $\mathbf{W}^{'}_{S}$, thus improving the representation power.  

However, these subject-specific models have unaligned latent spaces by default due to their different MLP weights. This makes face editing challenging since each subject's latent space has a different semantic meaning. Thus, we propose a solution to resolve this tradeoff between representation power and editability.
If any common attributes (\textit{e.g.}, expression) exist across different subjects, INFAMOUS-NeRF learns a single latent code for all subjects rather than many subject-specific latent codes, thus achieving semantically-aligned latent spaces. 
For example, all smiling subjects are modeled with the same shared expression latent code rather than one expression code per smiling subject. 
This constrains the hypernetwork to estimate subject-specific weights that can still leverage a single shared latent code, which maintains semantic alignment.  

 We train our model using multi-view images from the FaceScape dataset~\cite{yang2020facescape} with groundtruth pose, including extrinsic rigid transformation parameters ($\mathbf{R}_{i}$, $\mathbf{t}_{i}$) for $i=1,...,n$. Similar to standard NeRF training, we construct a ray from each sampled pixel ($x_{i,p}$, $y_{i,p}$) with the ray direction $\mathbf{d}_{i,p}$. 
We then compute the ray points $\mathbf{r}_{i, p, m}$ for $m=1,...,M$, where $M$ is the number of sampled points per ray, as: 
\begin{equation}
    \mathbf{r}_{i, p, m} = \mathbf{t}_{i}+(z_\text{near}+(m-1)\frac{z_\text{far}-z_\text{near}}{M-1})\mathbf{d}_{i,p},
\end{equation}
where $z_\text{near}$ and $z_\text{far}$ are the near and far bounds. Similar to MoFaNeRF, we have a separate appearance and shape MLP for predicting the RGB texture $\mathbf{c}$ and $\sigma$ respectively. Each query to the appearance MLP $F_{A}$ is then of the form $F_{A}(\mathbf{\alpha}, \mathbf{r}_{i, p, m}, \mathbf{d}_{i, p}; \mathbf{W}^{'}_{A}) \xrightarrow[]{}\mathbf{c}$ and each query to the shape MLP $F_{S}$ is of the form $F_{S}(\mathbf{\alpha}, \mathbf{\beta}, \mathbf{\epsilon}, \mathbf{r}_{i, p, m}, \mathbf{d}_{i, p}; \mathbf{W}^{'}_{S}) \xrightarrow[]{}\sigma$. We then render the pixel colors in the same manner as~\cite{NeRF}.

Beyond the standard NeRF formulation, we also introduce a novel constraint to improve the rendering quality along the face boundary by focusing on the surface point along each ray and enforcing photometric consistency. We discuss this constraint in Sec.~\ref{sec:psc}. 

For our second stage model, which we use as novel view refinement, we employ a conditional DDPM by adopting the method of~\cite{ilvr_adm}. Our DDPM has one spatial condition: the coarse novel view image $\mathbf{I}^{'}_{i}$ estimated by our stage $1$ NeRF model and thus has the following objective: 
\begin{equation}
 \Eb{t, \bx_0, \bepsilon}{ \left\| \bepsilon - \bepsilon_\theta(\sqrt{\bar\alpha_t} \bx_0 + \sqrt{1-\bar\alpha_t}\bepsilon, t, \mathbf{I}^{'}_{i}) \right\|^2}. \label{eq:training_objective_conditional_ours}
\end{equation} 
Our condition $\mathbf{I}^{'}_{i}$ is spatially concatenated with $\mathbf{x}_{t}$, where
 $\mathbf{x}_{t}= \sqrt{\bar\alpha_t}\bx_0 + \sqrt{1-\bar\alpha_t}\bepsilon$. 
To enable our final single-image face modeling method, we perform latent optimization to estimate $\mathbf{\theta}$, $\mathbf{\alpha}$, $\mathbf{\beta}$, and $\mathbf{\epsilon}$ given a single image, freezing all other parameters. See Fig.~\ref{fig:Architecture} for a detailed illustration of INFAMOUS-NeRF. 

\subsection{Photometric Surface Constraint}
\label{sec:psc}
To improve the rendering quality along the face boundary where the error is generally high, we add an additional photometric surface rendering constraint. The final rendering step in NeRF's volumetric rendering is to compute the pixel color  $\mathbf{C}=\sum\limits_{i=1}^{M}\omega_{i}\mathbf{c}_{i}$, where $\mathbf{c}_{i}$ is the predicted color of the $i^{th}$ ray point and $\omega_{i}$ is the volume density dependent weight. We choose the surface point $\mathbf{s}$ along each ray as the $k^{th}$ sampled point on the ray, where
\begin{align}
 k=\argmax_{j \in 1,2,....,M}\omega_{j} \label{eq:surfacePoint}
\end{align} 
and enforce that the predicted color $\mathbf{c}_{k}$ at point $\mathbf{s}$ is consistent with multi-view corresponding pixels, which are determined through projection. We represent this constraint as $\mathcal{L}_\text{sur}$, defined by 
\begin{equation}
\begin{split}
    \mathcal{L}_\text{sur} \!=\!\!\sum\limits_{p=1}^{N}\!(\|\hat{\mathbf{c}}_{1,p,k_{1,p}}\!-\mathbf{C}_{1, p}\|_\text{1}\!+\!\|\hat{\mathbf{c}}_{2,p,k_{2,p}}\!-\mathbf{C}_{2, p}\|_\text{1}\! \\ +\! 
    \|\hat{\mathbf{c}}_{1,p,k_{1,p}}\!-\mathbf{C}_{2, p_{1,2}}\|_\text{1}\!+\! \|\hat{\mathbf{c}}_{2,p,k_{2,p}}\!-\mathbf{C}_{1, p_{2,1}}\|_\text{1}),
\end{split}
\end{equation}
where $\hat{\mathbf{c}}_{i,p,k_{i,p}}$ is the color predicted at the $k_{i,p}^{th}$ ray point (the predicted surface point $\mathbf{s}_{i, p}$) for the $p^{th}$ pixel sampled from the $i^{th}$ image in the batch, $\mathbf{C}_{i, p}$ is the groundtruth color of the $p^{th}$ pixel in the $i^{th}$ batch image, and $\mathbf{C}_{i, p_{j,i}}$ is the groundtruth color of the pixel in the $i^{th}$ image corresponding to the surface point $\mathbf{s}_{j, p}$. 
This constraint enforces that the face boundary is learned well across all views due to its emphasis on the predicted surface color, thus improving the rendering. 

\subsection{Adaptive Sampling}
When sampling pixels to train NeRF methods, it is important to be conscious of the memory budget as each pixel involves casting a ray into the $3$D volume. Sampling every pixel in the image at every iteration is often infeasible given memory constraints, particularly at higher resolutions. Most NeRF methods select pixels by employing uniform sampling, which risks sacrificing fidelity due to undersampling regions with fine detail (\textit{e.g.}, the eyes, nose, and mouth) as well as oversampling nearly textureless regions such as the cheek. MoFaNeRF~\cite{zhuang2022mofanerf} proposes a landmark-based sampling scheme where $60\%$ of the pixels are sampled around the inner face landmarks and the rest are sampled uniformly. While this will naturally improve the quality of the rendered inner face, it ignores other areas where the error is generally high (\textit{e.g.}, the head boundary). In addition, uniform sampling is more redundant if some regions (\textit{e.g.}, the background) are textureless or uniform.  

In light of this, we introduce a novel adaptive sampling method that stochastically selects new pixels at each iteration based on their loss values. Our intuition is that a higher loss generally indicates that a pixel lies within a region with large gradients or fine details and should be sampled more frequently to properly maintain the fidelity. We thus want to assign certain pixels a higher probability of being sampled than others by producing an $H\!\!\times\!\!W$ probability map $\mathbf{P}_{i}$ for each image $\mathbf{I}_{i}$ that contains the per-pixel sampling probability. We describe our sampling strategy in detail in the supplementary materials.

During training, we sample $50\%$ of the rays around the inner face landmarks and the other $50\%$ using our adaptive sampling scheme. We find that this partition yields the best of both worlds: INFAMOUS-NeRF renders the inner face with high-fidelity and adaptively assigns significant weight to other regions with high reconstruction error not covered by the landmarks such as the head boundary.  

\subsection{Loss Functions}
In stage $1$, our method is supervised at the image level by a reconstruction loss $\mathcal{L}_\text{recon}$, a structural dissimilarity loss $\mathcal{L}_\text{DSSIM}$ similar to ~\cite{PhysicsGuidedRelighting},  
and a gradient loss $\mathcal{L}_\text{gradient}$ to ensure that image-level gradients are preserved in the rendered images, which we describe in detail in the supplementary materials. 


To improve the rendering quality near the face boundary, we also add our novel photometric surface rendering constraint $\mathcal{L}_\text{sur}.$ 
Our final training loss for stage $1$ is therefore: 
\begin{equation}
\begin{split}
    \mathcal{L}_\text{total}=\alpha_{1}\mathcal{L}_\text{recon}+\alpha_{2}\mathcal{L}_
    \text{DSSIM}+ \alpha_{3}\mathcal{L}_\text{gradient}+\alpha_{4}\mathcal{L}_\text{sur},
\end{split} 
\end{equation}
where $\alpha_{1}=\alpha_{2}=\alpha_{3}=\alpha_{4}=1$ are the loss weights. 
As mentioned before, the training objective for our stage $2$ conditional DDPM is given by ~Eq.~\ref{eq:training_objective_conditional_ours}. 

%% file: sec/4_experiments.tex
\section{Experiments}

\begin{table*}[t!]

\caption{\small
\textbf{Representation Power}. We compare our $3$DMM fitting and novel view synthesis on the FaceScape test set with face modeling methods. Our model achieves the highest PSNR and SSIM (mean $\pm$ standard deviation) by a large margin, indicating that INFAMOUS-NeRF is able to substantially improve the representation power. We compare with MoFaNeRF~\cite{zhuang2022mofanerf} using $56$ test subjects evaluating on the full head and neck and HeadNeRF~\cite{hong2021headnerf} using $41$ subjects evaluating only on the rendered face and thus report two separate sets of results. 
}\label{tab:3DMMFittingNVSEvaluation}
\vspace{-2mm}
\begin{center}
\scalebox{0.75}{
\setlength\tabcolsep{8pt}  
\begin{tabular}{c c c c c}
\hline
Method & PSNR (Fit) & SSIM (Fit) & PSNR (NVS) & SSIM (NVS) \\
\hline
MoFaNeRF~\cite{zhuang2022mofanerf} & $17.00\pm1.71$ & $0.6672\pm0.0528$ & $15.18\pm1.87$ & $0.7505\pm0.0830$ \\
\hline
INFAMOUS-NeRF (Ours) & $\mathbf{20.81\pm1.18}$ & $\mathbf{0.8943\pm0.0350}$ & $\mathbf{19.15\pm1.80}$ & $\mathbf{0.8589\pm0.0437}$\\
\hline
\hline
HeadNeRF~\cite{hong2021headnerf} & $18.47\pm3.15$ & $0.8959\pm0.0500$ & $17.12\pm2.83$ & $0.8376\pm0.0606$ \\
\hline
INFAMOUS-NeRF (Ours) & $\mathbf{24.33\pm1.46}$ & $\mathbf{0.9426\pm0.0178}$ & $\mathbf{22.43\pm1.94}$ & $\mathbf{0.9186\pm0.0302}$\\
\hline
\end{tabular}}
\end{center}

\end{table*}

\begin{figure*}[t]
\vspace{-2mm}
\begin{center}

\begin{minipage}[t]{0.102\linewidth}
\centering
\includegraphics[width=\linewidth]{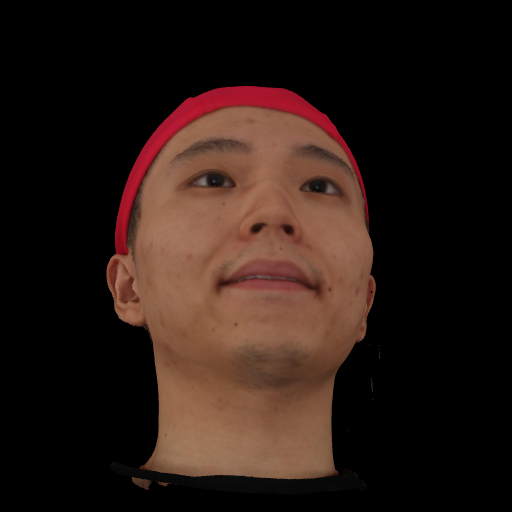}
\end{minipage}
\begin{minipage}[t]{0.102\linewidth}
\centering
\includegraphics[width=\linewidth]{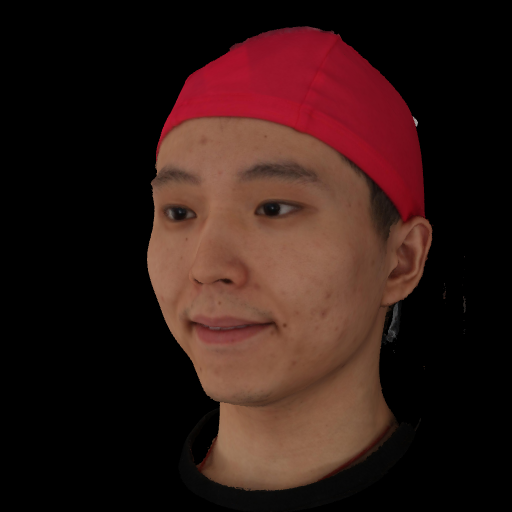}
\end{minipage}
\begin{minipage}[t]{0.102\linewidth}
\centering
\includegraphics[width=\linewidth]{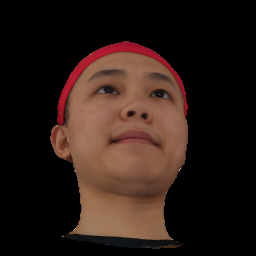}
\end{minipage}
\begin{minipage}[t]{0.102\linewidth}
\centering
\includegraphics[width=\linewidth]{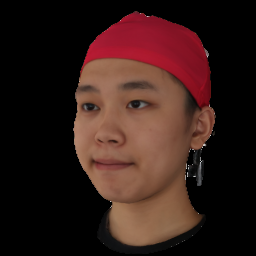}
\end{minipage}
\begin{minipage}[t]{0.102\linewidth}
\centering
\includegraphics[width=\linewidth]{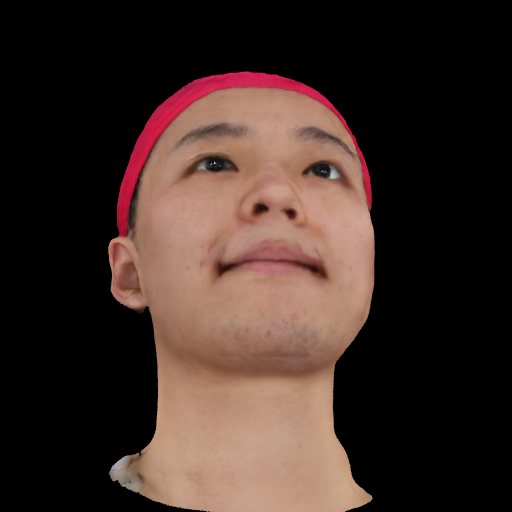}
\end{minipage}
\begin{minipage}[t]{0.102\linewidth}
\centering
\includegraphics[width=\linewidth]{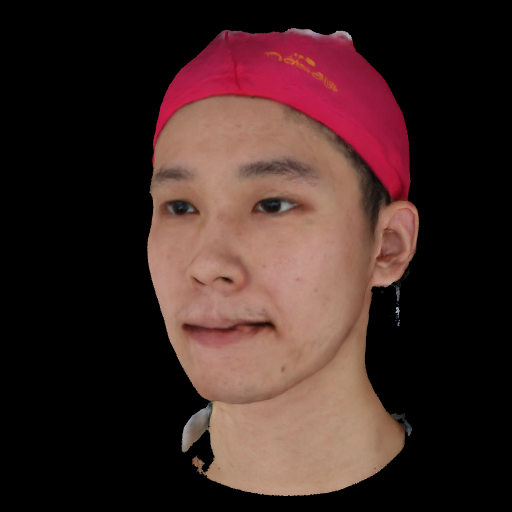}
\end{minipage}
\begin{minipage}[t]{0.102\linewidth}
\centering
\includegraphics[width=\linewidth]{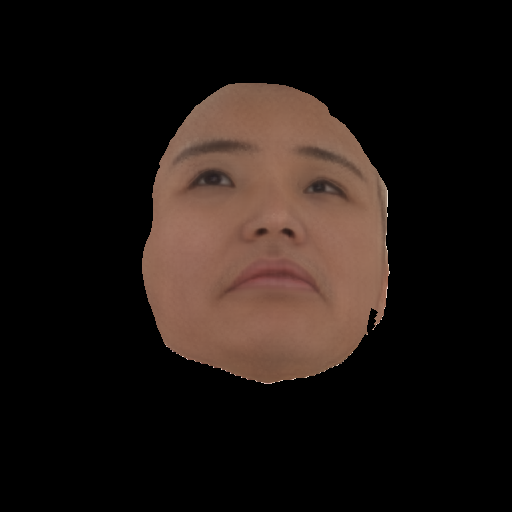} \\
\end{minipage}
\begin{minipage}[t]{0.102\linewidth}
\centering
\includegraphics[width=\linewidth]{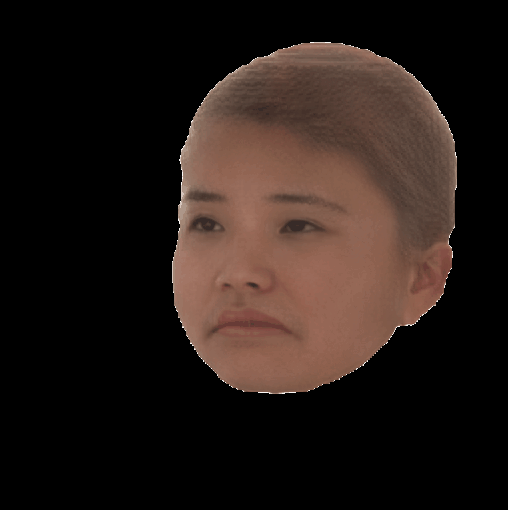} \\
\end{minipage}

\begin{minipage}[t]{0.102\linewidth}
\centering
\includegraphics[width=\linewidth]{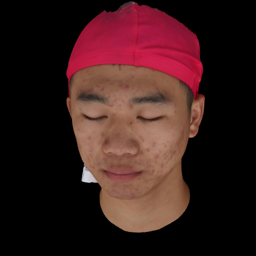}
\end{minipage}
\begin{minipage}[t]{0.102\linewidth}
\centering
\includegraphics[width=\linewidth]{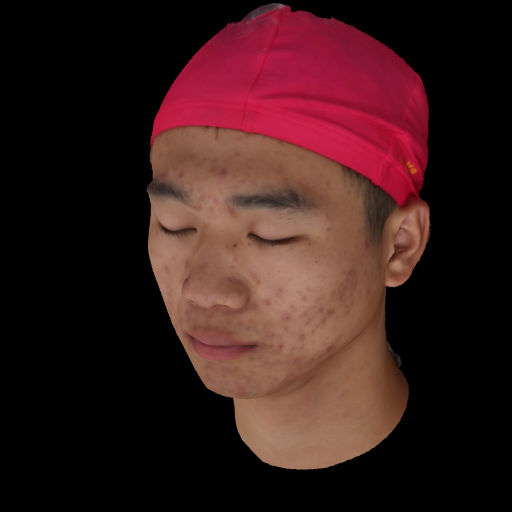}
\end{minipage}
\begin{minipage}[t]{0.102\linewidth}
\centering
\includegraphics[width=\linewidth]{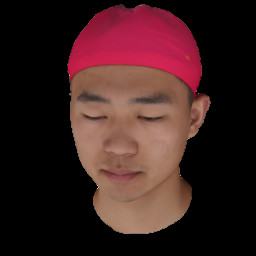}
\end{minipage}
\begin{minipage}[t]{0.102\linewidth}
\centering
\includegraphics[width=\linewidth]{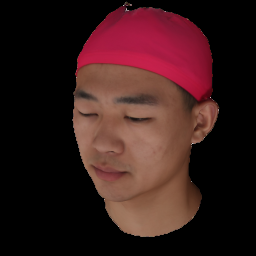}
\end{minipage}
\begin{minipage}[t]{0.102\linewidth}
\centering
\includegraphics[width=\linewidth]{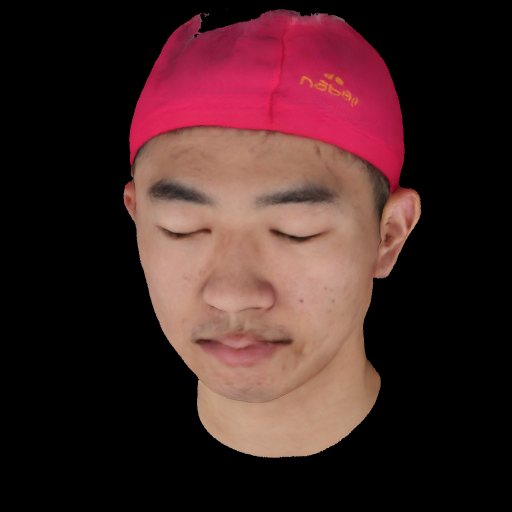}
\end{minipage}
\begin{minipage}[t]{0.102\linewidth}
\centering
\includegraphics[width=\linewidth]{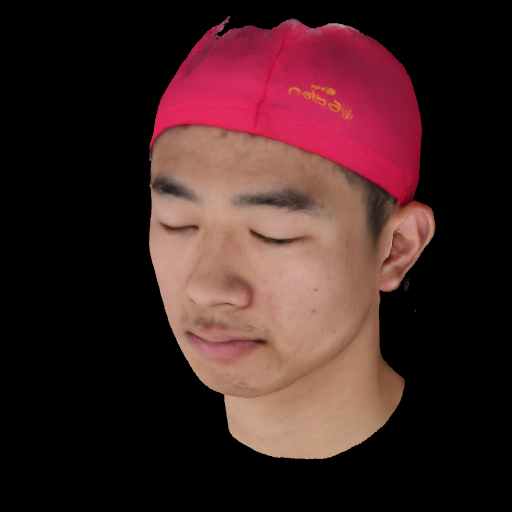}
\end{minipage}
\begin{minipage}[t]{0.102\linewidth}
\centering
\includegraphics[width=\linewidth]{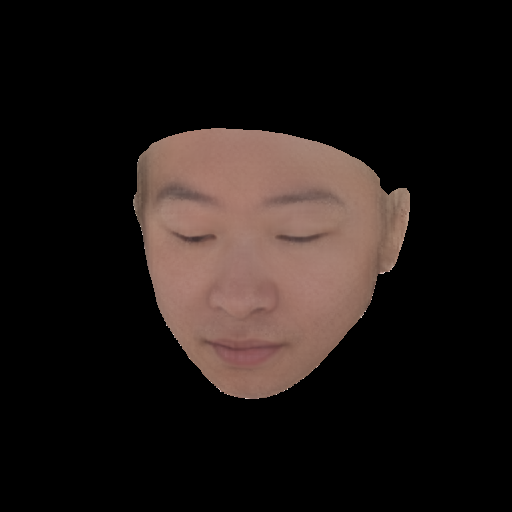} \\
\end{minipage}
\begin{minipage}[t]{0.102\linewidth}
\centering
\includegraphics[width=\linewidth]{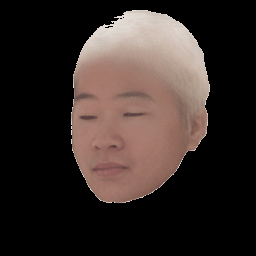} \\
\end{minipage}


\begin{minipage}[t]{0.102\linewidth}
\centering
\includegraphics[width=\linewidth]{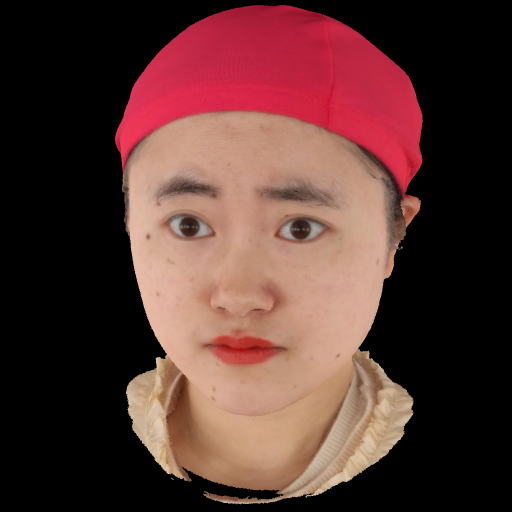} \\
\end{minipage}
\begin{minipage}[t]{0.102\linewidth}
\centering
\includegraphics[width=\linewidth]{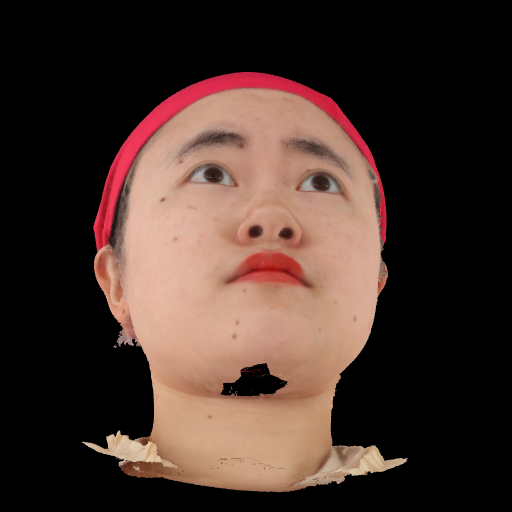} \\
\end{minipage}
\begin{minipage}[t]{0.102\linewidth}
\centering
\includegraphics[width=\linewidth]{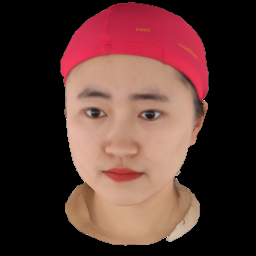} \\
\end{minipage}
\begin{minipage}[t]{0.102\linewidth}
\centering
\includegraphics[width=\linewidth]{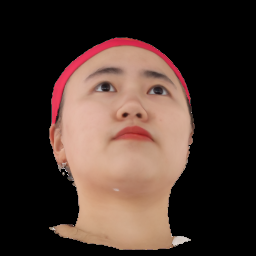} \\
\end{minipage}
\begin{minipage}[t]{0.102\linewidth}
\centering
\includegraphics[width=\linewidth]{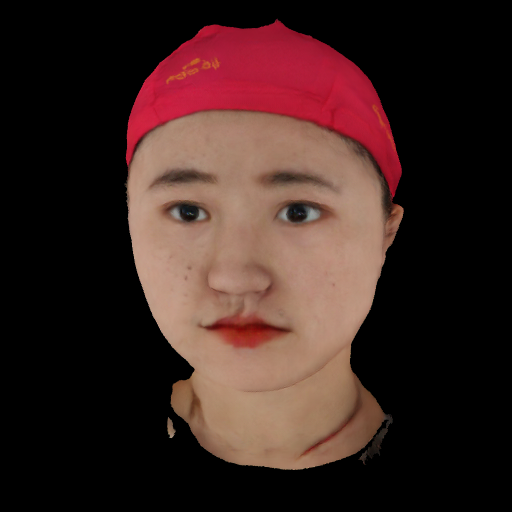} \\
\end{minipage}
\begin{minipage}[t]{0.102\linewidth}
\centering
\includegraphics[width=\linewidth]{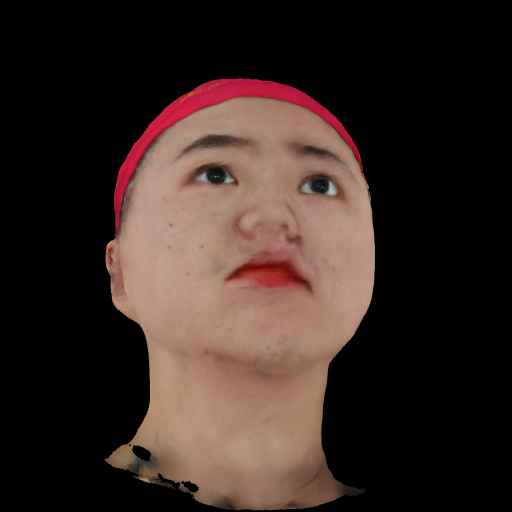} \\
\end{minipage}
\begin{minipage}[t]{0.102\linewidth}
\centering
\includegraphics[width=\linewidth]{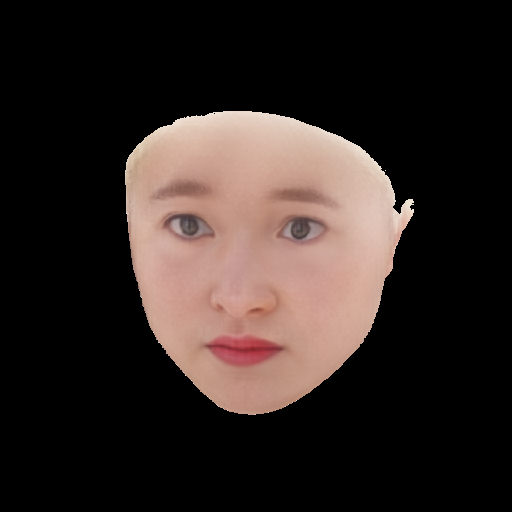} \\
\end{minipage}
\begin{minipage}[t]{0.102\linewidth}
\centering
\includegraphics[width=\linewidth]{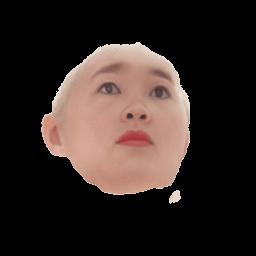} \\
\end{minipage}

\begin{minipage}[t]{0.102\linewidth}
\centering
\includegraphics[width=\linewidth]{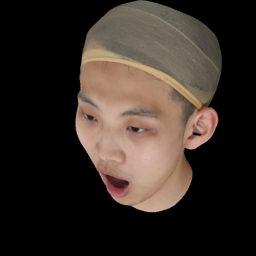} \\
\small a) Input Image
\end{minipage}
\begin{minipage}[t]{0.102\linewidth}
\centering
\includegraphics[width=\linewidth]{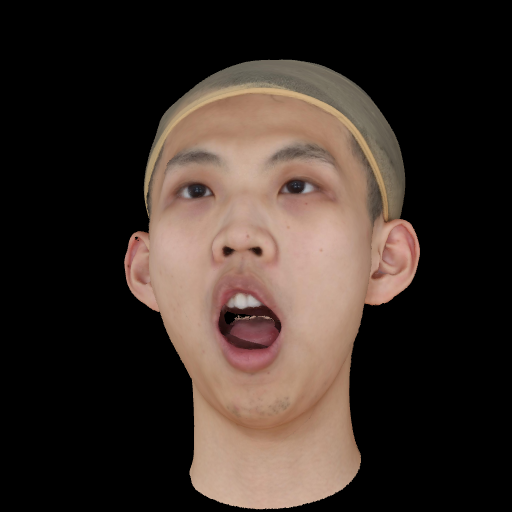} \\
\small b) Target Image
\end{minipage}
\begin{minipage}[t]{0.102\linewidth}
\centering
\includegraphics[width=\linewidth]{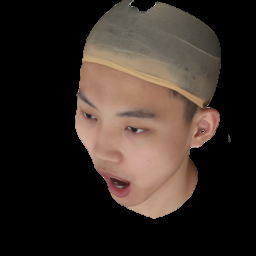} \\
\small c) Proposed (Fit)
\end{minipage}
\begin{minipage}[t]{0.102\linewidth}
\centering
\includegraphics[width=\linewidth]{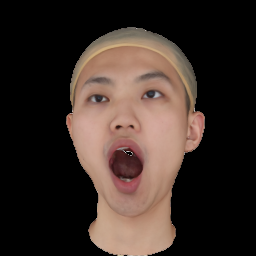} \\
\small d) Proposed (NVS)
\end{minipage}
\begin{minipage}[t]{0.102\linewidth}
\centering
\includegraphics[width=\linewidth]{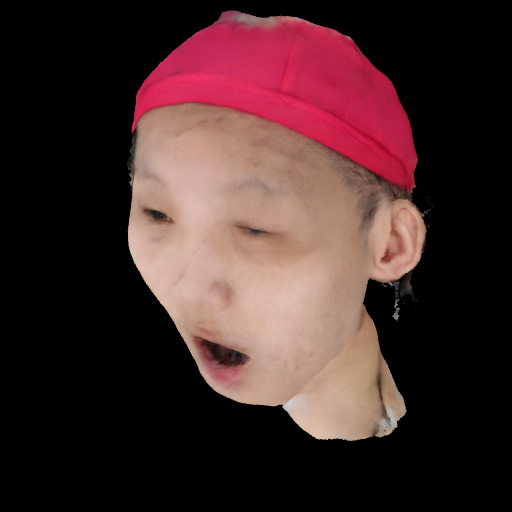} \\
\small e) MoFa~\cite{zhuang2022mofanerf} (Fit)
\end{minipage}
\begin{minipage}[t]{0.102\linewidth}
\centering
\includegraphics[width=\linewidth]{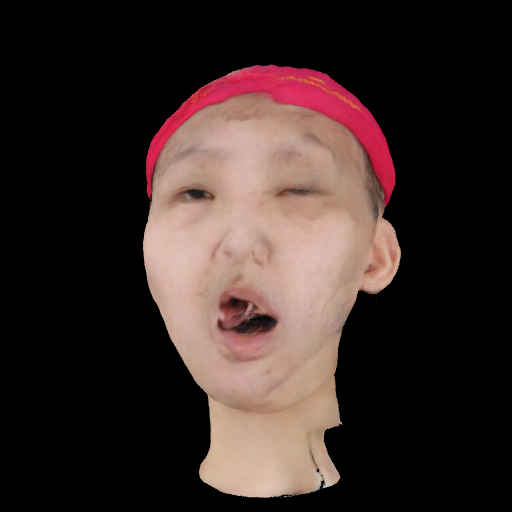} \\
\small f) MoFa~\cite{zhuang2022mofanerf} (NVS)
\end{minipage}
\begin{minipage}[t]{0.102\linewidth}
\centering
\includegraphics[width=\linewidth]{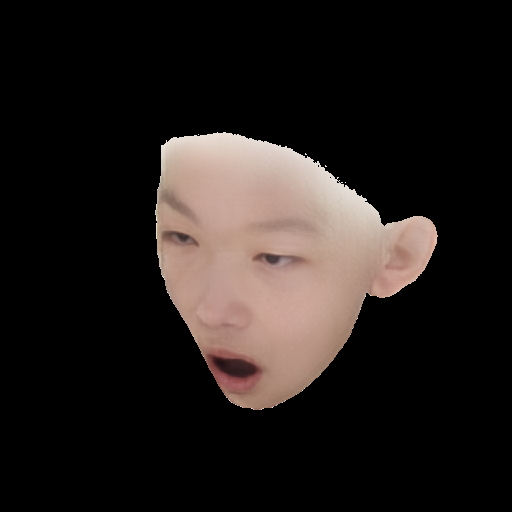} \\
\small g) Head~\cite{hong2021headnerf} (Fit)
\end{minipage}
\begin{minipage}[t]{0.102\linewidth}
\centering
\includegraphics[width=\linewidth]{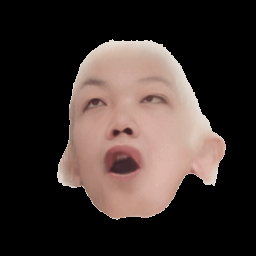} \\
\small h) Head~\cite{hong2021headnerf} (NVS)
\end{minipage}
\vspace{-2mm}
\caption{\small \textbf{Qualitative Representation Power on FaceScape}. We demonstrate qualitative results on $3$DMM fitting (Fit) and novel view synthesis (NVS) compared with the baselines. INFAMOUS-NeRF noticeably represents the testing subjects more faithfully based on skin tone, facial structures, and expressions. Here we \textit {avoid cherry-picking} our best results and instead show the subject with the highest PSNR of $23.17$ ($1^{st}$ row), the subject with the median PSNR of $21.02$ ($2^{nd}$ row), and the subject with the lowest PSNR of $17.83$ ($3^{rd}$ row). The $4^{th}$ row shows the most challenging expression in the FaceScape dataset: "mouth stretch". 
}\label{fig:RepresentationPower}
\end{center}\vspace{-10mm}
\end{figure*}

Similar to MoFaNeRF~\cite{zhuang2022mofanerf}, we train our stage $1$ NeRF model using the $300$ training subjects from the FaceScape dataset~\cite{yang2020facescape}. 
Since each subject in FaceScape contains images of $20$ different expressions, we semantically align the expression latent space by learning a $20\times64$ set of expression latent codes that are shared across all training subjects, one per unique expression. 
This enforces that the learned expression latent space possesses the same semantic meaning across all subjects. To train the stage $2$ conditional DDPM, we use latent optimization to learn the optimal $\mathbf{\theta}$, $\mathbf{\alpha}$, $\mathbf{\beta}$, and $\mathbf{\epsilon}$ for training images and then render coarse predicted images with paired groundtruth. 
\vspace{-5mm}






\begin{figure}
\begin{center}
\includegraphics[width=0.8\linewidth]{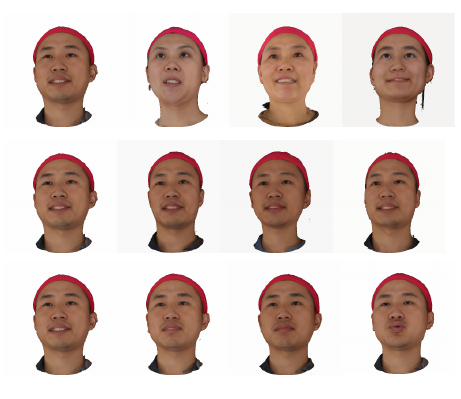}
\vspace{-5mm}
\caption{\small \textbf{Face Editing}. We perform latent code swapping between different subjects (top: appearance, middle: shape, bottom: expression). Our method disentangles these components well and generates diverse identities and expressions.  
}\label{fig:GenericFaceEditing}
\vspace{-10mm}
\end{center}
\end{figure}

\paragraph{Representation Power.} To evaluate INFAMOUS-NeRF's representation power, we perform novel view synthesis and $3$DMM fitting experiments against face modeling baselines that can also edit expressions, appearance, and shape~\cite{zhuang2022mofanerf, hong2021headnerf}. 
We use the remaining $56$ unseen subjects in FaceScape~\cite{yang2020facescape} when comparing against MoFaNeRF~\cite{zhuang2022mofanerf} and $40$ of these $56$ subjects when comparing against HeadNeRF~\cite{hong2021headnerf} since the other $16$ subjects are in the HeadNeRF training set. We sample $1$ random input view per subject and $3$ random target views, all of which are novel poses that are unseen in training. 
Each of the $56$ subjects has one of the $20$ expressions in FaceScape, randomly selected. This leads to a total test set size of $168$ images for evaluating novel view synthesis and $56$ images for evaluating $3$DMM fitting ($120$ and $41$ respectively for HeadNeRF). 
We achieve SoTA novel view synthesis and $3$DMM fitting performance (See Tab.~\ref{tab:3DMMFittingNVSEvaluation}), demonstrating our superior representation power. 
We compute the mean PSNR and SSIM in the full portrait mask of the groundtruth image when comparing with MoFaNeRF and in the face region only when comparing with HeadNeRF to be fair to each baseline. Qualitatively, we faithfully represent each subject and are noticeably superior to the baselines (See Fig.~\ref{fig:RepresentationPower}). 
We can represent diverse skin tones, facial structures, and expressions much more accurately due to higher representation power from our hypernetwork. 
MoFaNeRF also does not converge to a reasonable solution in some cases due to instability from training  with adversarial loss. 
INFAMOUS-NeRF instead treats the refinement stage as a separate step, which ensures that the coarse image from stage $1$ will stably converge. HeadNeRF tends to produce blurry results and sometimes does not faithfully represent the subject (\textit{e.g.}, skin tones and eye colors). 
\vspace{-6mm}

\begin{figure*}
\vspace{-2mm}
\begin{center}

\begin{minipage}[t]{0.102\linewidth}
\centering
\includegraphics[width=\linewidth]{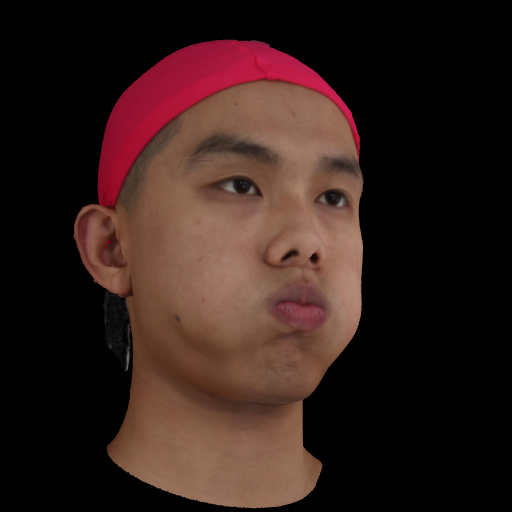}
\end{minipage}
\begin{minipage}[t]{0.102\linewidth}
\centering
\includegraphics[width=\linewidth]{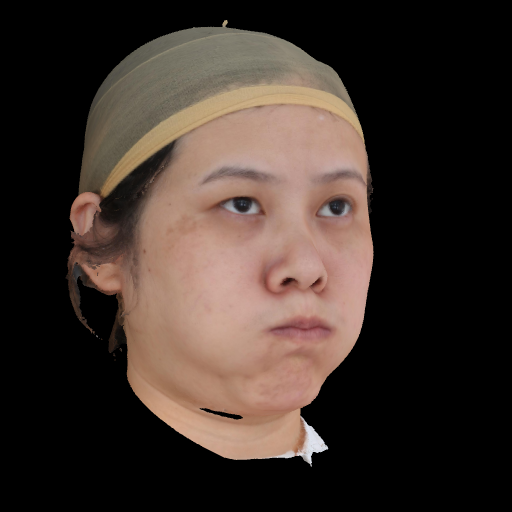}
\end{minipage}
\begin{minipage}[t]{0.102\linewidth}
\centering
\includegraphics[width=\linewidth]{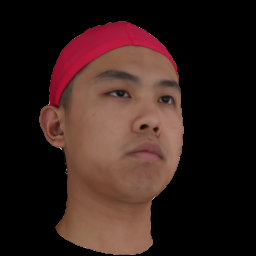}
\end{minipage}
\begin{minipage}[t]{0.102\linewidth}
\centering
\includegraphics[width=\linewidth]{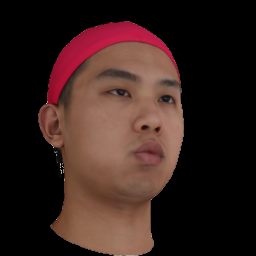}
\end{minipage}
\begin{minipage}[t]{0.102\linewidth}
\centering
\includegraphics[width=\linewidth]{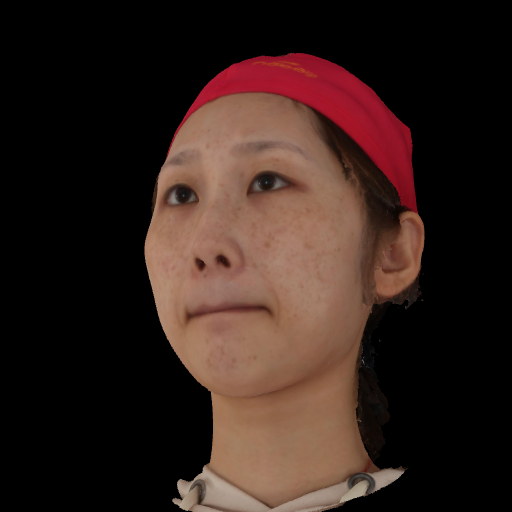}
\end{minipage}
\begin{minipage}[t]{0.102\linewidth}
\centering
\includegraphics[width=\linewidth]{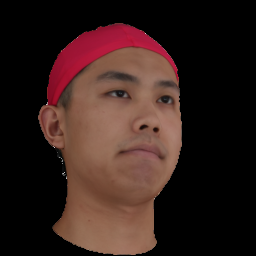}
\end{minipage}
\begin{minipage}[t]{0.102\linewidth}
\centering
\includegraphics[width=\linewidth]{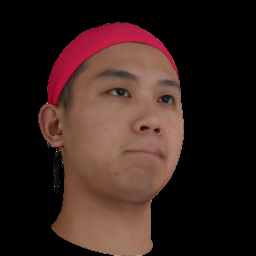}
\end{minipage}

\begin{minipage}[t]{0.102\linewidth}
\centering
\includegraphics[width=\linewidth]{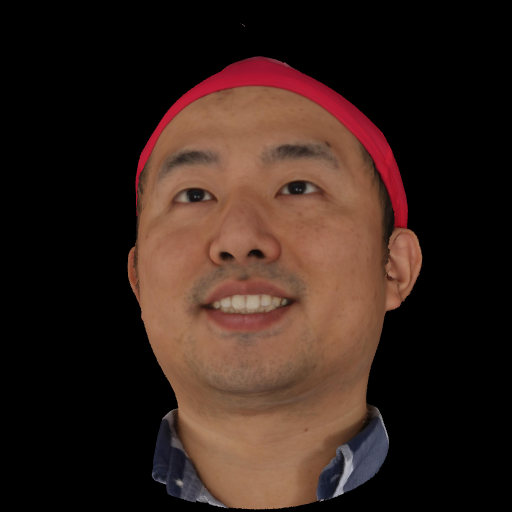} \\
\small a) Source
\end{minipage}
\begin{minipage}[t]{0.102\linewidth}
\centering
\includegraphics[width=\linewidth]{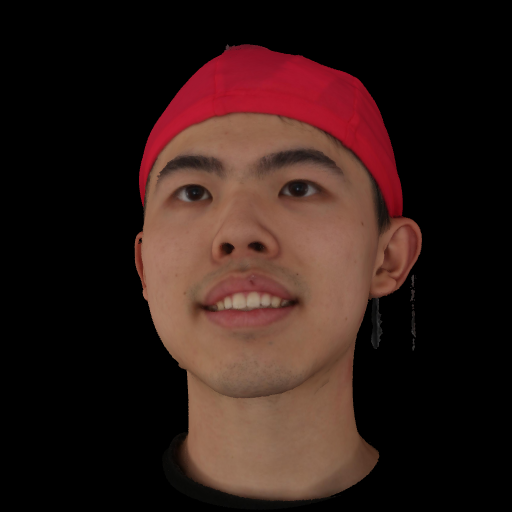} \\
\small b) Target
\end{minipage}
\begin{minipage}[t]{0.102\linewidth}
\centering
\includegraphics[width=\linewidth]{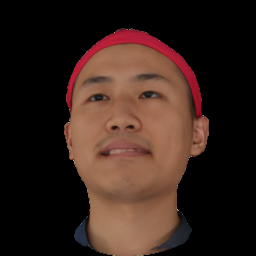} \\
\small c) Naive HN
\end{minipage}
\begin{minipage}[t]{0.102\linewidth}
\centering
\includegraphics[width=\linewidth]{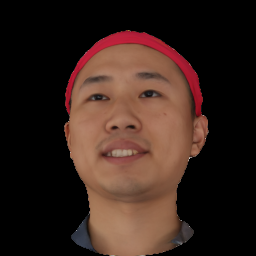} \\
\small d) Proposed 
\end{minipage}
\begin{minipage}[t]{0.102\linewidth}
\centering
\includegraphics[width=\linewidth]{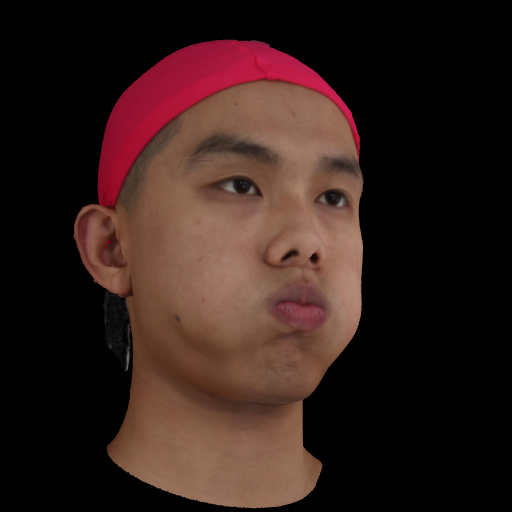} \\
\small e) Target
\end{minipage}
\begin{minipage}[t]{0.102\linewidth}
\centering
\includegraphics[width=\linewidth]{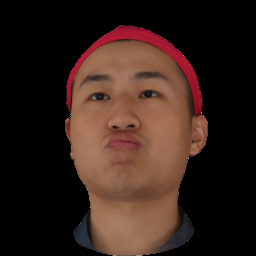} \\
\small f) Naive HN
\end{minipage}
\begin{minipage}[t]{0.102\linewidth}
\centering
\includegraphics[width=\linewidth]{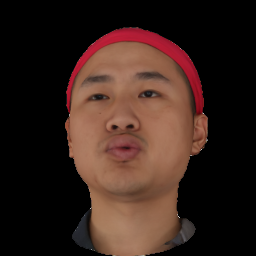} \\
\small g) Proposed
\end{minipage}
\vspace{-2mm}
\caption{\small \textbf{Semantic Alignment of Expression}. By applying the target's expression code to the source, INFAMOUS-NeRF properly transfers the target's expression, showing the expression latent space is semantically aligned. The first target in each row has the same expression and the source maintains its original expression, whereas the second target has a different expression that transfers to the source.
}\label{fig:SemanticAlignment}
\end{center}\vspace{-6mm}
\end{figure*}

\paragraph{Semantically-Aligned Face Editing.} To show that INFAMOUS-NeRF resolves the classic hypernetwork tradeoff between representation power and editability, we conduct expression transfer experiments between source and target image pairs. By applying the target's learned expression code to the source, INFAMOUS-NeRF can transfer the target's expression to the source (See Fig.~\ref{fig:SemanticAlignment}). This transferability between different subjects shows that the expression latent space is semantically aligned. Notably, INFAMOUS-NeRF can learn this semantic alignment from scratch instead of requiring a pretrained model like prior work~\cite{hyperstyle,hyperinverter}. We further train a second model that employs a more naive hypernetwork training strategy, where each subject has their own unique expression codes (the $20\!\!\times\!\!64$ expression table becomes $300\!\!\times\!\!20\!\!\times\!\!64$).
Note that this "Naive HN" model cannot properly perform expression transfer, showing that our training strategy is necessary to achieve semantic alignment. 
For example, "Naive HN" cannot maintain the "cheek blowing" expression or transfer over the "lip roll" expression in the first row and instead produces frowns in both cases. We also show results on appearance and shape editing, which involves swapping $\mathbf{\alpha}$/$\mathbf{W}^{'}_{A}$ and $\mathbf{\beta}$/$\mathbf{W}^{'}_{S}$ respectively between different subjects. As seen in Fig.~\ref{fig:GenericFaceEditing}, INFAMOUS-NeRF disentangles appearance, shape, and expression well and can generate diverse identities and expressions.  
\vspace{-5mm}


\begin{table*}
\caption{\small
\textbf{In-the-Wild Fitting.} We evaluate our fitting performance on $80$ FFHQ test images and $26$ CelebAHQ images. Our method achieves significantly higher PSNR/SSIM than the baselines, showing that its higher representation power extends to in-the-wild images. 
}\label{tab:FFHQFitting}
\vspace{-5mm}
\begin{center}
\scalebox{0.75}{
\setlength\tabcolsep{8pt}  
\begin{tabular}{c c c c c}
\hline
Method & PSNR (FFHQ) & SSIM (FFHQ) & PSNR (CelebAHQ) & SSIM (CelebAHQ) \\
\hline
MoFaNeRF~\cite{zhuang2022mofanerf} & $14.32\pm1.61$ & $0.5180\pm0.0918$ & $13.90\pm2.46$ & $0.5054\pm0.0764$ \\
\hline
INFAMOUS-NeRF (Ours) & $\mathbf{19.07\pm2.11}$ & $\mathbf{0.7847\pm0.0890}$ & $\mathbf{19.36\pm1.18}$ & $\mathbf{0.7710\pm0.0803}$\\
\hline
\hline
HeadNeRF~\cite{hong2021headnerf} & $14.22\pm4.23$ & $0.7124\pm0.1170$ & $13.13\pm4.78$ & $0.6568\pm0.1301$\\
\hline
INFAMOUS-NeRF (Ours) & $\mathbf{20.65\pm2.21}$ & $\mathbf{0.8032\pm0.0869}$ & $\mathbf{19.97\pm1.21}$ & $\mathbf{0.7645\pm0.0892}$ \\
\hline
\end{tabular}}
\end{center}

\vspace{-5mm}
\end{table*}

\begin{table*}[hbt!]
\caption{\small
\textbf{Identity Preservation}. We compute the average AdaFace cosine similarity in the FaceScape test set as well as FFHQ and CelebAHQ to evaluate our method's identity preservation.
INFAMOUS-NeRF is noticeably superior in identity preservation with the highest average AdaFace cosine similarity (CS) for both fitting and novel view synthesis (mean $\pm$ standard deviation). 
}\label{tab:AdaFaceCS}
\vspace{-5mm}
\begin{center}
\scalebox{0.75}{
\setlength\tabcolsep{8pt}  
\begin{tabular}{c c c c c}
\hline
Method & CS (Fit-FaceScape) & CS (NVS-FaceScape) & CS (Fit-FFHQ) & CS (Fit-CelebAHQ) \\
\hline
MoFaNeRF~\cite{zhuang2022mofanerf} & $0.4089\pm0.1302$ & $0.3227\pm0.1252$ & $0.1288\pm0.1011$ & $0.1013\pm0.0738$\\
\hline
Ours & $\mathbf{0.4452\pm0.0942}$ & $\mathbf{0.3592\pm0.0955}$ & $\mathbf{0.2255\pm0.0876}$ & $\mathbf{0.2173\pm0.0859}$ \\
\hline
\hline
HeadNeRF~\cite{hong2021headnerf} & $0.3027\pm0.1012$ & $0.2149\pm0.1188$ & $0.2164\pm0.1111$ & $0.1934\pm0.0997$\\
\hline
Ours & $\mathbf{0.4482\pm0.0973}$ & $\mathbf{0.3579\pm0.1017}$ & $\mathbf{0.2221\pm0.0881}$ & $\mathbf{0.2119\pm0.0860}$ \\
\hline
\end{tabular}}
\end{center}

\vspace{-5mm}
\end{table*}

\begin{figure}[hbt!]
\vspace{-2mm}
\begin{center}

\begin{minipage}[t]{0.212\linewidth}
\centering
\includegraphics[width=\linewidth]{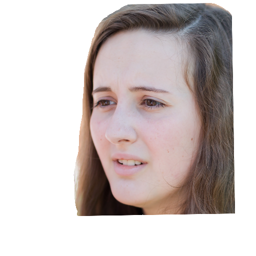}
\end{minipage}
\begin{minipage}[t]{0.212\linewidth}
\centering
\includegraphics[width=\linewidth]{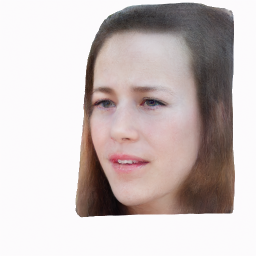}
\end{minipage}
\begin{minipage}[t]{0.212\linewidth}
\centering
\includegraphics[width=\linewidth]{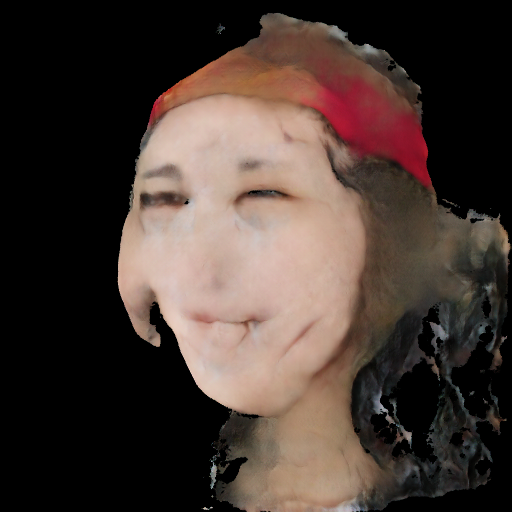}
\end{minipage}
\begin{minipage}[t]{0.212\linewidth}
\centering
\includegraphics[width=\linewidth]{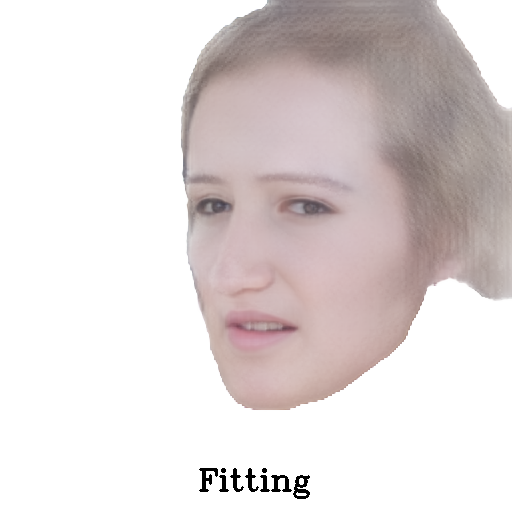}
\end{minipage}



\begin{minipage}[t]{0.212\linewidth}
\centering
\includegraphics[width=\linewidth]{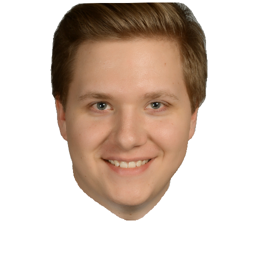} \\
\small a) Target
\end{minipage}
\begin{minipage}[t]{0.212\linewidth}
\centering
\includegraphics[width=\linewidth]{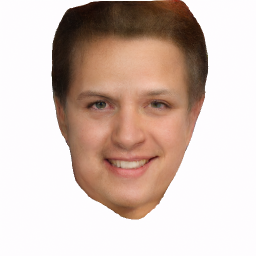} \\
\small b) Proposed
\end{minipage}
\begin{minipage}[t]{0.212\linewidth}
\centering
\includegraphics[width=\linewidth]{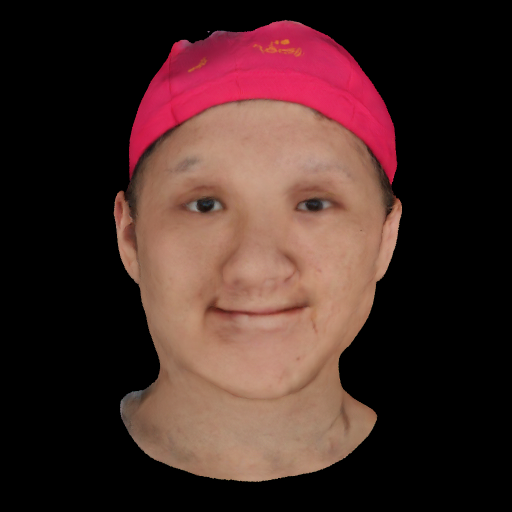} \\
\small c) MoFa~\cite{zhuang2022mofanerf}
\end{minipage}
\begin{minipage}[t]{0.212\linewidth}
\centering
\includegraphics[width=\linewidth]{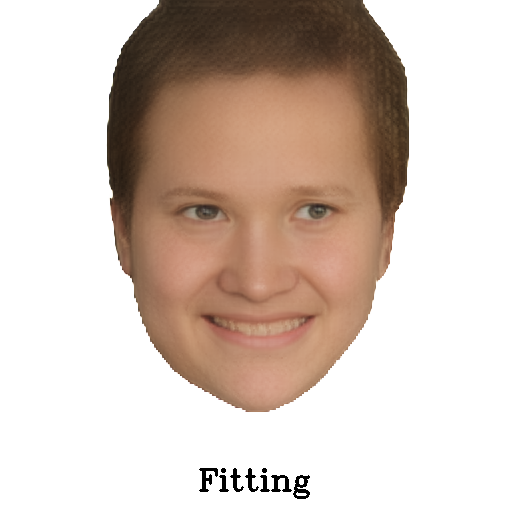} \\
\small d) Head~\cite{hong2021headnerf}
\end{minipage}
\vspace{-0.8mm}
\caption{\small \textbf{FFHQ Representation Power}. Compared to the baselines, our method is able to substantially improve the representation power and faithfulness to the subject identity for in-the-wild subjects, especially different hairstyles and skin tones.
}\label{fig:FFHQRepresentationPower}
\end{center}\vspace{-10mm}
\end{figure}

\begin{figure}[hbt!]
\vspace{-2mm}
\begin{center}

\begin{minipage}[t]{0.212\linewidth}
\centering
\includegraphics[width=\linewidth]{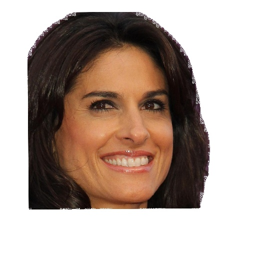}
\end{minipage}
\begin{minipage}[t]{0.212\linewidth}
\centering
\includegraphics[width=\linewidth]{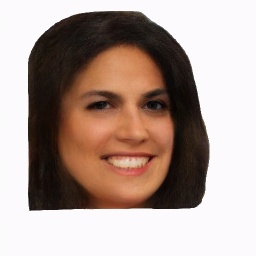}
\end{minipage}
\begin{minipage}[t]{0.212\linewidth}
\centering
\includegraphics[width=\linewidth]{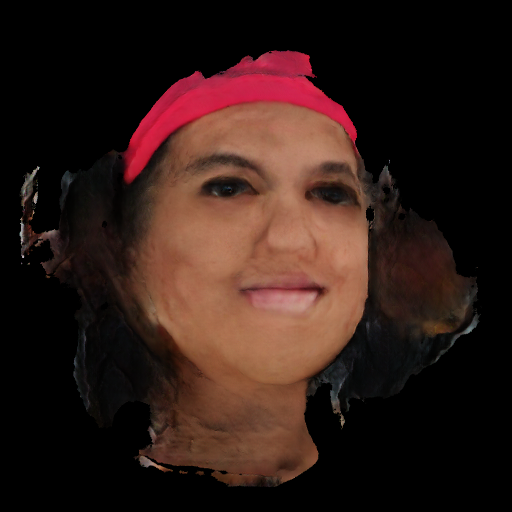}
\end{minipage}
\begin{minipage}[t]{0.212\linewidth}
\centering
\includegraphics[width=\linewidth]{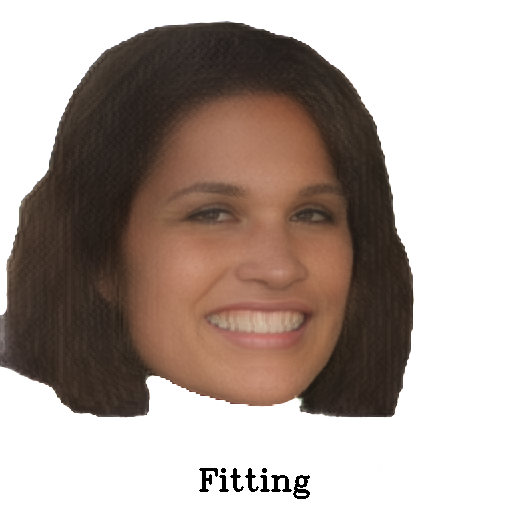}
\end{minipage}

\begin{minipage}[t]{0.212\linewidth}
\centering
\includegraphics[width=\linewidth]{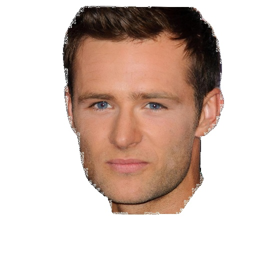} \\
\small a) Target
\end{minipage}
\begin{minipage}[t]{0.212\linewidth}
\centering
\includegraphics[width=\linewidth]{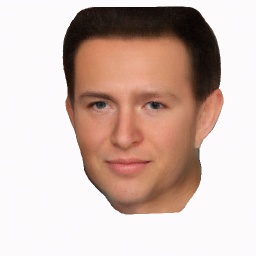} \\
\small b) Proposed
\end{minipage}
\begin{minipage}[t]{0.212\linewidth}
\centering
\includegraphics[width=\linewidth]{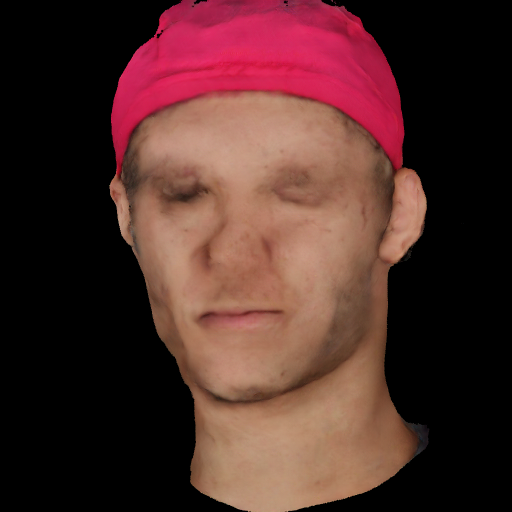} \\
\small c) MoFa~\cite{zhuang2022mofanerf}
\end{minipage}
\begin{minipage}[t]{0.212\linewidth}
\centering
\includegraphics[width=\linewidth]{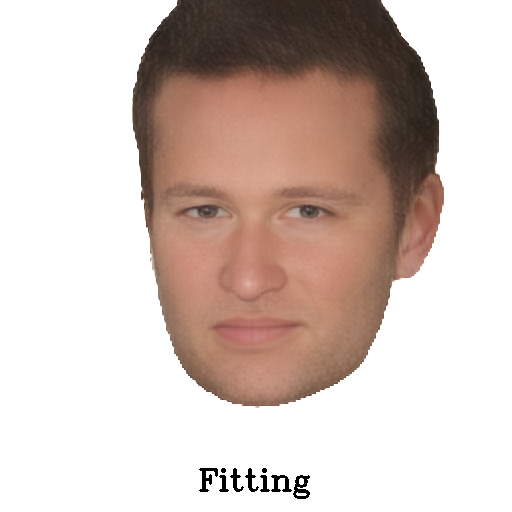} \\
\small d) Head~\cite{hong2021headnerf}
\end{minipage}


\vspace{-0.8mm}
\caption{\small \textbf{CelebAHQ Representation Power}. We show results on CelebAHQ, which neither our method nor our baselines have trained on. Similar to FFHQ, our model demonstrates significantly higher faithfulness to the test subjects compared to the baselines.  
}\label{fig:CelebAHQRepresentationPower}
\end{center}\vspace{-9mm}
\end{figure}

\paragraph{In-the-Wild Performance}

To improve INFAMOUS-NeRF's generalizability to in-the-wild images, we finetune our FaceScape model on $2386$ FFHQ~\cite{FFHQ} images with $6$ different expressions: smile, grin, neutral, sadness, anger, and mouth stretch. We use the FFHQ expression labels provided by~\cite{FFHQ_attributes}, which are generated using a network that estimates facial attributes. We still encourage semantic alignment by enforcing a single shared latent code for each of the $6$ expressions and share the existing corresponding expression codes from FaceScape. For example, every smiling subject in our FFHQ finetuning set will use the smiling latent code for the FaceScape subjects. 

To evaluate our fitting performance on in-the-wild images, we further assemble a test set of $80$ diverse FFHQ subjects with all $6$ expressions well represented in the test set as well as a second test set of $26$ CelebAHQ~\cite{CelebA-HQ} subjects. As seen in Tab.~\ref{tab:FFHQFitting}, we substantially outperform both baselines in fitting performance thanks to our higher representation power from hypernetwork modeling. This shows that our method achieves higher representation power than our baselines not only on the FaceScape dataset, but also for in-the-wild images. Notably, HeadNeRF trains on over $4000$ FFHQ images and yet still achieves significantly worse PSNR/SSIM than our method fine tuned on only around $2000$ images. Qualitatively, as seen in Figs.~\ref{fig:FFHQRepresentationPower} and~\ref{fig:CelebAHQRepresentationPower}, our fitting is much more faithful to the subject than MoFaNeRF and HeadNeRF. MoFaNeRF often cannot converge to a good solution or generalize well to in-the-wild subjects and also generates a red cap instead of hair on the crown of the head. HeadNeRF performs better than MoFaNeRF for in-the-wild images, but often produces blurry or faded results and is sometimes unfaithful to the identity (\textit{e.g.}, hairstyle/hair color, gaze direction, skin tone). 

\begin{figure*}[t]
\vspace{-2mm}
\begin{center}

\begin{minipage}[t]{0.12\linewidth}
\centering
\includegraphics[width=\linewidth]{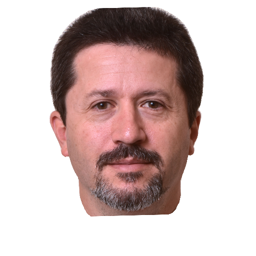} \\
\end{minipage}
\begin{minipage}[t]{0.12\linewidth}
\centering
\includegraphics[width=\linewidth]{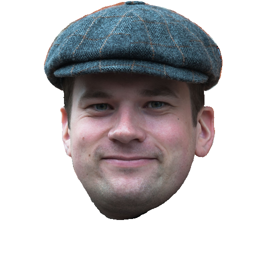} \\
\end{minipage}
\begin{minipage}[t]{0.12\linewidth}
\centering
\includegraphics[width=\linewidth]{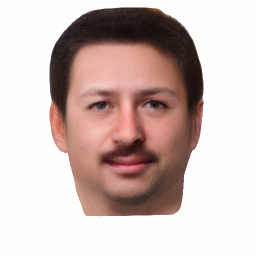} \\
\end{minipage}
\begin{minipage}[t]{0.12\linewidth}
\centering
\includegraphics[width=\linewidth]{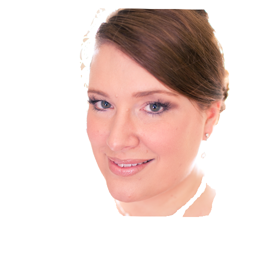} \\
\end{minipage}
\begin{minipage}[t]{0.12\linewidth}
\centering
\includegraphics[width=\linewidth]{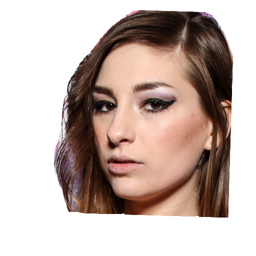} \\
\end{minipage}
\begin{minipage}[t]{0.12\linewidth}
\centering
\includegraphics[width=\linewidth]{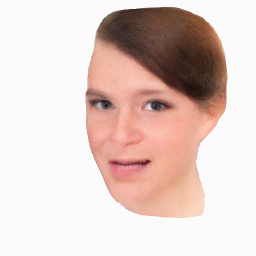} \\
\end{minipage}

\begin{minipage}[t]{0.12\linewidth}
\centering
\includegraphics[width=\linewidth]{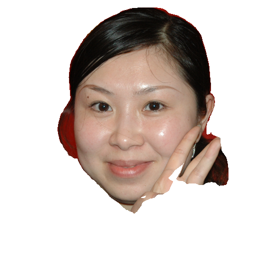} \\
\small a) Source
\end{minipage}
\begin{minipage}[t]{0.12\linewidth}
\centering
\includegraphics[width=\linewidth]{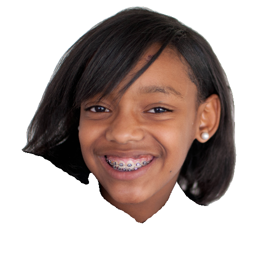} \\
\small b) Target
\end{minipage}
\begin{minipage}[t]{0.12\linewidth}
\centering
\includegraphics[width=\linewidth]{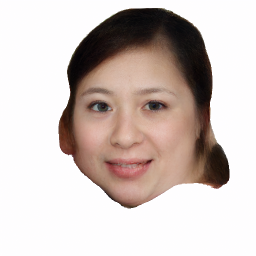} \\
\small c) Proposed
\end{minipage}
\begin{minipage}[t]{0.12\linewidth}
\centering
\includegraphics[width=\linewidth]{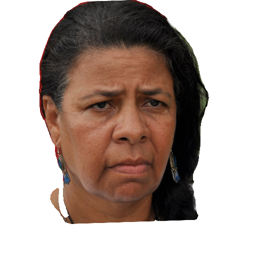} \\
\small d) Source
\end{minipage}
\begin{minipage}[t]{0.12\linewidth}
\centering
\includegraphics[width=\linewidth]{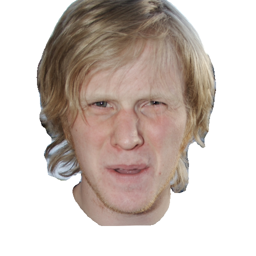} \\
\small e) Target
\end{minipage}
\begin{minipage}[t]{0.12\linewidth}
\centering
\includegraphics[width=\linewidth]{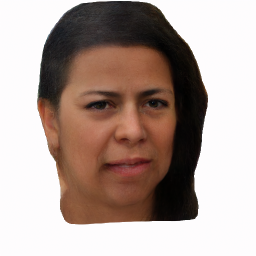} \\
\small f) Proposed
\end{minipage}
\vspace{-2mm}
\caption{\small \textbf{FFHQ Expression Semantic Alignment}. We show expression transfer on in-the-wild images with a semantically aligned expression space for editing. (Top left: neutral to smile, top right: grin to neutral, bottom left: smile to grin, bottom right: sadness to anger)
}\label{fig:FFHQSemanticAlignment}
\end{center}\vspace{-6mm}
\end{figure*}

We also demonstrate that we can achieve semantic alignment of the expression latent space on in-the-wild images, as shown in Fig.~\ref{fig:FFHQSemanticAlignment}. We are able to transfer diverse expressions such as smile, grin, neutral, and sadness which shows that our expression latent space is semantically aligned between different subjects and suitable for expression editing. 

\vspace{-5mm}


\begin{table*}
\caption{\small
\textbf{Ablations}. We ablate our hypernetwork modeling, $\mathcal{L}_\text{sur}$, and adaptive sampling and show that they all improve the representation power. As our contributions are all centered around improving NeRF, we ablate by reporting stage $1$ results (w/o diffusion).   
}\label{tab:Ablations}
\vspace{-5mm}
\begin{center}
\scalebox{0.75}{
\setlength\tabcolsep{5pt}  
\begin{tabular}{c c c c c}
\hline
Method & PSNR (Fit) & SSIM (Fit) & PSNR (NVS) & SSIM (NVS) \\
\hline
Ours (w/o $\mathcal{L}_\text{sur}$) & $20.77\pm0.8557$ & $0.9016\pm0.0336$ & $18.53\pm1.5312$ & $0.8567\pm0.0455$ \\
\hline
Ours (w/o adaptive sampling) & $20.95\pm1.0801$ & $0.9034\pm1.0.0319$ & $18.57\pm1.7267$ & $0.8579\pm0.0431$ \\
\hline
Ours (w/o hypernetwork) & $20.05\pm1.0903$ & $0.8863\pm0.0401$ & $18.32\pm1.7892$ & $0.8523\pm0.0461$ \\
\hline
Proposed & $\mathbf{21.00\pm0.9610}$ & $\mathbf{0.9039\pm0.0321}$ & $\mathbf{18.73\pm1.7136}$ & $\mathbf{0.8600\pm0.0433}$ \\
\hline
\end{tabular}}
\end{center}

\vspace{-6mm}
\end{table*}

\paragraph{Identity Preservation}
To evaluate how well we are able to preserve subject identity both in fitting and during novel view synthesis, we use the average AdaFace~\cite{kim2022adaface} feature cosine similarity across all FaceScape~\cite{yang2020facescape} test subjects as a metric. Specifically, we compute the AdaFace cosine similarity between each test image and its corresponding input image and compare our identity preservation with HeadNeRF~\cite{hong2021headnerf} and MoFaNeRF~\cite{zhuang2022mofanerf}. Besides FaceScape, we also compare our identity preservation on in-the-wild images by evaluating on our FFHQ and CelebAHQ test sets. As seen in Tab.~\ref{tab:AdaFaceCS}, our identity preservation is superior to both baselines, largely owing to our higher representation power from subject-specific hypernetwork modeling. 

\vspace{-5mm}


\paragraph{Ablation Studies.}

To ablate our contributions to representation power, we train a separate model that learns a single set of weights for all subjects rather than using hypernetworks.
We further ablate our contribution to improving semantic alignment by training a model that learns subject-specific appearance, shape, and expression codes. We also ablate our photometric surface rendering constraint by training our stage $1$ NeRF model without $\mathcal{L}_\text{sur}$. 
Finally, we ablate our adaptive sampling algorithm by training stage $1$ with half landmark and half uniform sampling. 
As shown in Tab.~\ref{tab:Ablations}, using hypernetworks significantly improves the representation power over learning a single model. $\mathcal{L}_\text{sur}$ is able to improve the representation power thanks to its focus on properly rendering the face boundary. Our adaptive sampling strategy is also able to improve the overall PSNR and SSIM over purely landmark and uniform sampling due to focusing on high error regions such as the head boundary. 

%% file: sec/5_conclusion.tex
\section{Conclusion}
We  propose INFAMOUS-NeRF: a face modeling method that improves representation power through the use of hypernetworks to estimate subject-specific NeRF MLPs and also resolves the tradeoff between representation power and editability without requiring a large pretrained model. 
Through expression transfer experiments, we show that our learned latent spaces are semantically aligned despite the subject-specific MLPs having different weights. We demonstrate quantitatively and qualitatively that we achieve SoTA performance on novel view synthesis and $3$DMM fitting on FaceScape and in-the-wild images. We hope that our method will inspire future work aiming to increase $3$D face modeling representation power.
\vspace{-4mm}
\paragraph{Limitations.} While our model does benefit from the expression diversity of FaceScape~\cite{yang2020facescape} as well as in-the-wild images from the FFHQ~\cite{FFHQ} dataset, it can sometimes struggle to handle out-of-distribution expressions. While FFHQ does capture the diversity of in-the-wild expressions, it is still somewhat biased and contains mostly smiles and grins. One way to resolve this issue is to include training images from other datasets that contain expressions that are not well represented in FaceScape and FFHQ. Another limitation is the expensive initial optimization required to regress $\theta$, $\alpha$, $\beta$, and $\epsilon$ for a testing image, which is also a limitation of MoFaNeRF~\cite{zhuang2022mofanerf}. A solution to this could be to explore how to train a separate model that can quickly map from an input image to its corresponding optimized latent codes to bypass this expensive optimization step. 
\vspace{-3mm}
\paragraph{Potential Societal Impacts} 

Face modeling is an old problem that has received consistent interest from the vision and graphics communities, especially in recent years. While it carries ethical concerns, society as a whole has largely agreed that it is worth exploring given its widespread applications so long as biometric policies and licenses are followed. Like most face modeling methods, INFAMOUS-NeRF can potentially be used to generate deepfake images by altering appearance, shape, and expression. We certainly do not condone or encourage this as INFAMOUS-NeRF is intended for positive applications such as generating facial avatars in AR/VR. We hope that deepfake methods can continue to advance and potentially use our edited images as additional training data for deepfake detection. 

%% file: sec/X_suppl.tex
\clearpage
\setcounter{page}{1}
\maketitlesupplementary


\section{Additional In-the-Wild Results}

\begin{table*}
\caption{\small
\textbf{DDPM Ablations}. We demonstrate that the improvement in representation power is coming from our hypernetwork modeling, not the DDPM. The DDPM's purpose is simply to improve the image quality, as seen by the fact that it does not consistently improve the quantitative metrics. 
}\label{tab:AblationsDDPM}
\vspace{-5mm}
\begin{center}
\setlength\tabcolsep{5pt}  
\begin{tabular}{c c c c c}
\hline
Method & PSNR (Fit) & SSIM (Fit) & PSNR (NVS) & SSIM (NVS) \\
\hline
Ours (w/o DDPM) & $\mathbf{21.00\pm0.9610}$ & $\mathbf{0.9039\pm0.0321}$ & $18.73\pm1.7136$ & $\mathbf{0.8600\pm0.0433}$ \\
\hline
Ours (w/o hypernetwork) & $20.05\pm1.0903$ & $0.8863\pm0.0401$ & $18.32\pm1.7892$ & $0.8523\pm0.0461$ \\
\hline
Proposed & $20.81\pm1.18$ & $0.8943\pm0.0350$ & $\mathbf{19.15\pm1.80}$ & $0.8589\pm0.0437$ \\
\hline
\end{tabular}
\end{center}
\vspace{-6mm}
\end{table*}

\begin{figure*}[hbt!]
\vspace{-2mm}
\begin{center}

\begin{minipage}[t]{0.212\linewidth}
\centering
\includegraphics[width=\linewidth]{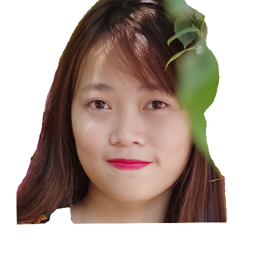}
\end{minipage}
\begin{minipage}[t]{0.212\linewidth}
\centering
\includegraphics[width=\linewidth]{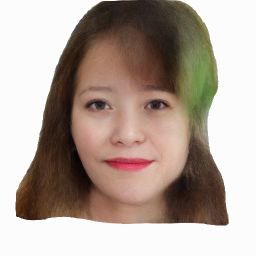}
\end{minipage}
\begin{minipage}[t]{0.212\linewidth}
\centering
\includegraphics[width=\linewidth]{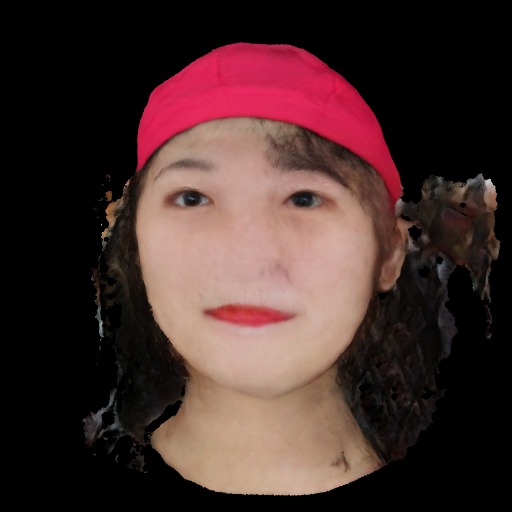}
\end{minipage}
\begin{minipage}[t]{0.212\linewidth}
\centering
\includegraphics[width=\linewidth]{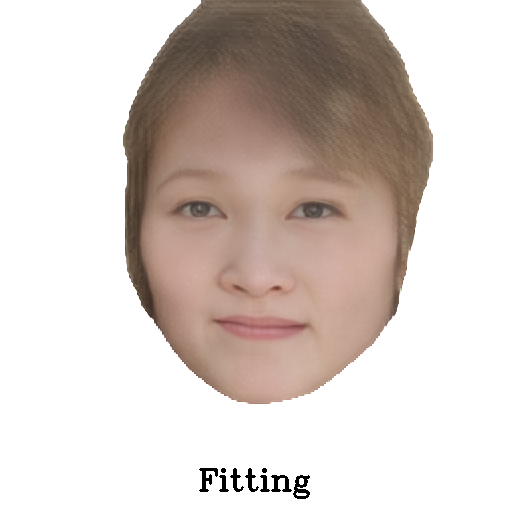}
\end{minipage}

\begin{minipage}[t]{0.212\linewidth}
\centering
\includegraphics[width=\linewidth]{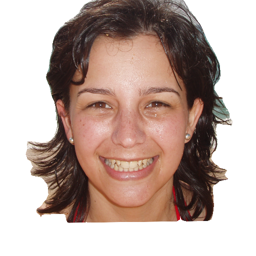}
\end{minipage}
\begin{minipage}[t]{0.212\linewidth}
\centering
\includegraphics[width=\linewidth]{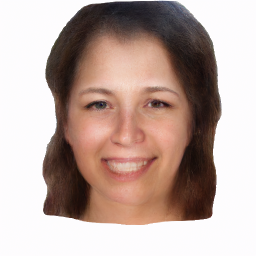}
\end{minipage}
\begin{minipage}[t]{0.212\linewidth}
\centering
\includegraphics[width=\linewidth]{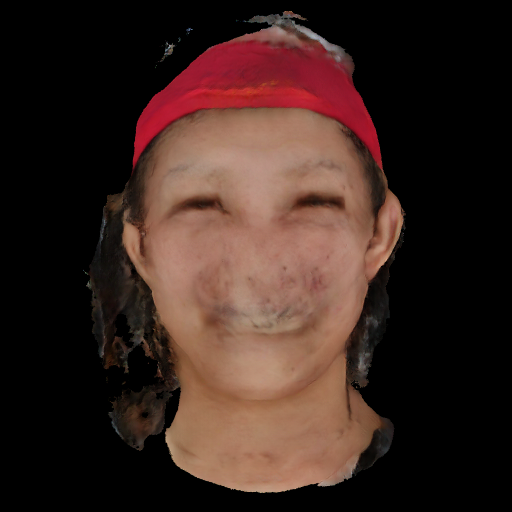}
\end{minipage}
\begin{minipage}[t]{0.212\linewidth}
\centering
\includegraphics[width=\linewidth]{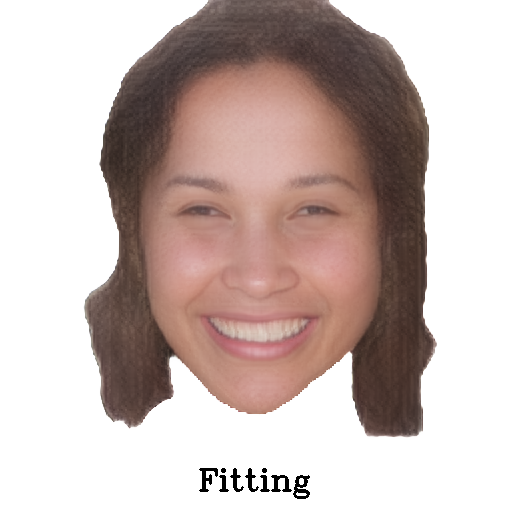}
\end{minipage}

\begin{minipage}[t]{0.212\linewidth}
\centering
\includegraphics[width=\linewidth]{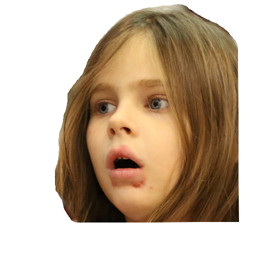}
\end{minipage}
\begin{minipage}[t]{0.212\linewidth}
\centering
\includegraphics[width=\linewidth]{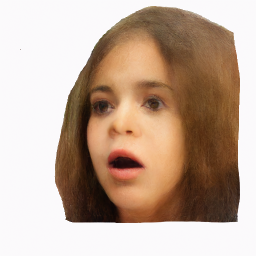}
\end{minipage}
\begin{minipage}[t]{0.212\linewidth}
\centering
\includegraphics[width=\linewidth]{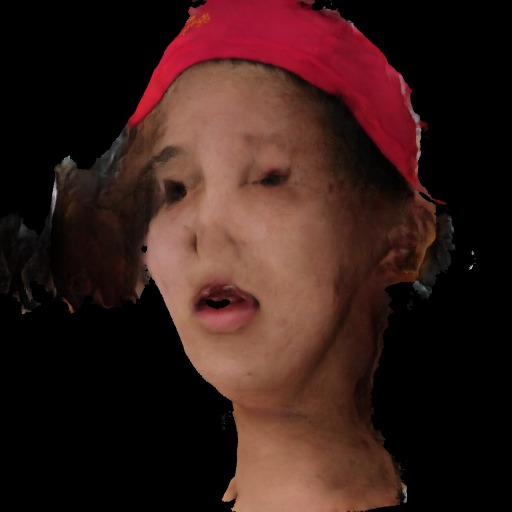}
\end{minipage}
\begin{minipage}[t]{0.212\linewidth}
\centering
\includegraphics[width=\linewidth]{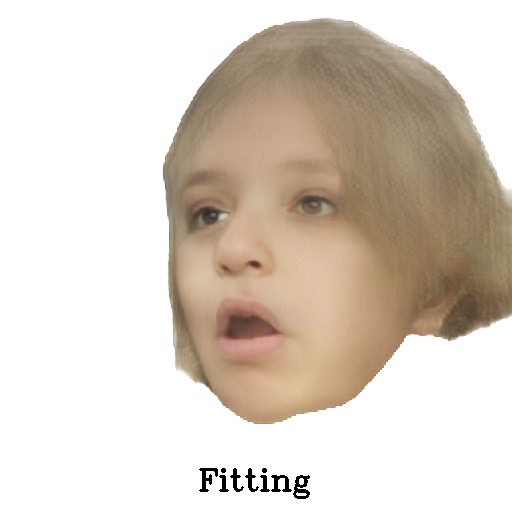}
\end{minipage}


\begin{minipage}[t]{0.212\linewidth}
\centering
\includegraphics[width=\linewidth]{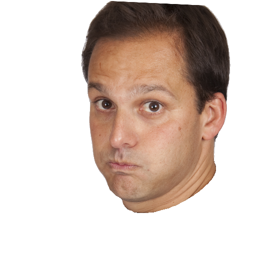}
\end{minipage}
\begin{minipage}[t]{0.212\linewidth}
\centering
\includegraphics[width=\linewidth]{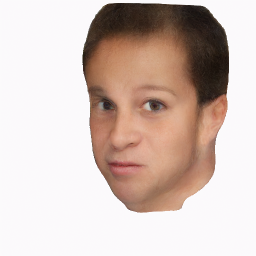}
\end{minipage}
\begin{minipage}[t]{0.212\linewidth}
\centering
\includegraphics[width=\linewidth]{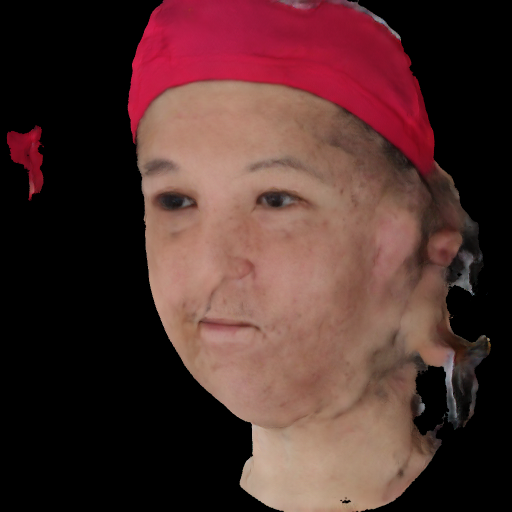}
\end{minipage}
\begin{minipage}[t]{0.212\linewidth}
\centering
\includegraphics[width=\linewidth]{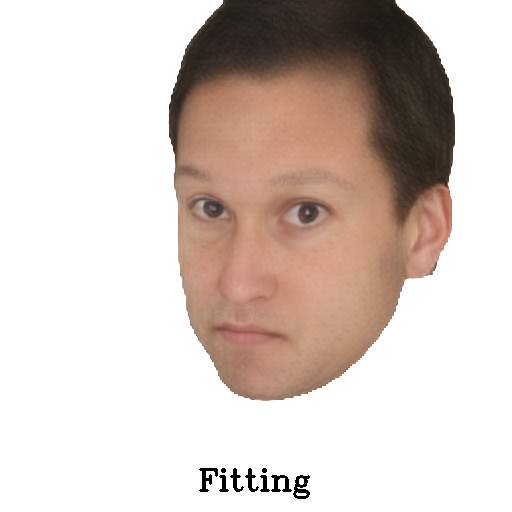}
\end{minipage}

\begin{minipage}[t]{0.212\linewidth}
\centering
\includegraphics[width=\linewidth]{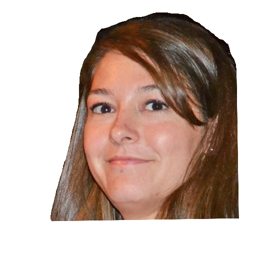} \\
\small a) Target
\end{minipage}
\begin{minipage}[t]{0.212\linewidth}
\centering
\includegraphics[width=\linewidth]{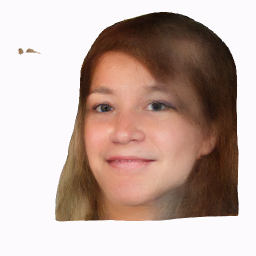} \\
\small b) Proposed
\end{minipage}
\begin{minipage}[t]{0.212\linewidth}
\centering
\includegraphics[width=\linewidth]{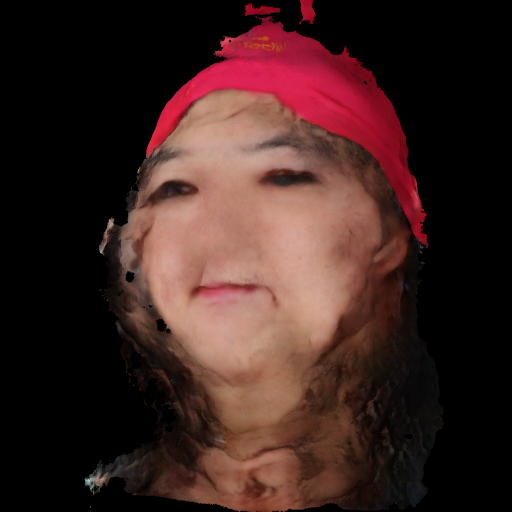}  \\
\small c) MoFaNeRF~\cite{zhuang2022mofanerf}
\end{minipage}
\begin{minipage}[t]{0.212\linewidth}
\centering
\includegraphics[width=\linewidth]{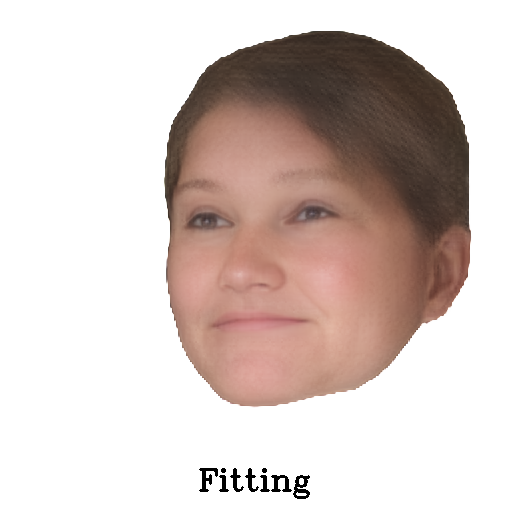} \\
\small d) HeadNeRF~\cite{hong2021headnerf}
\end{minipage}

\vspace{-0.8mm}
\caption{\small \textbf{FFHQ Representation Power}. Compared to the baselines, our method is able to substantially improve the representation power and faithfulness to the subject identity for in-the-wild subjects. In particular, we're able to properly represent different hairstyles and skin tones which is a problem for both baselines. 
}\label{fig:FFHQRepresentationPower}
\end{center}\vspace{-4mm}
\end{figure*}

\begin{figure*}[hbt!]
\vspace{-2mm}
\begin{center}


\begin{minipage}[t]{0.212\linewidth}
\centering
\includegraphics[width=\linewidth]{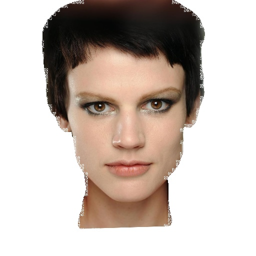}
\end{minipage}
\begin{minipage}[t]{0.212\linewidth}
\centering
\includegraphics[width=\linewidth]{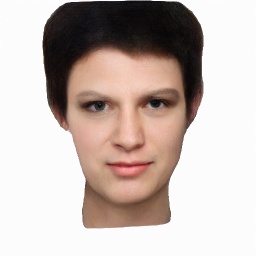}
\end{minipage}
\begin{minipage}[t]{0.212\linewidth}
\centering
\includegraphics[width=\linewidth]{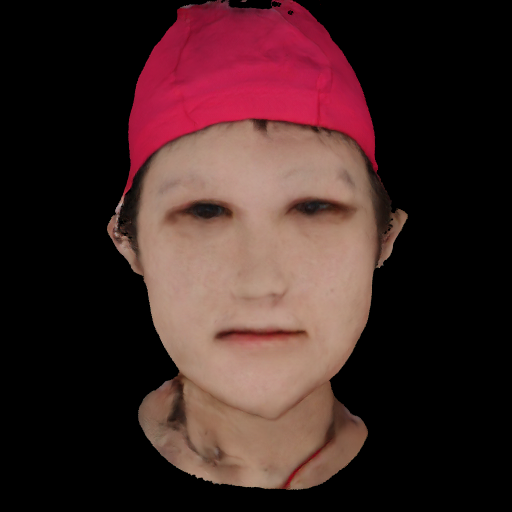}
\end{minipage}
\begin{minipage}[t]{0.212\linewidth}
\centering
\includegraphics[width=\linewidth]{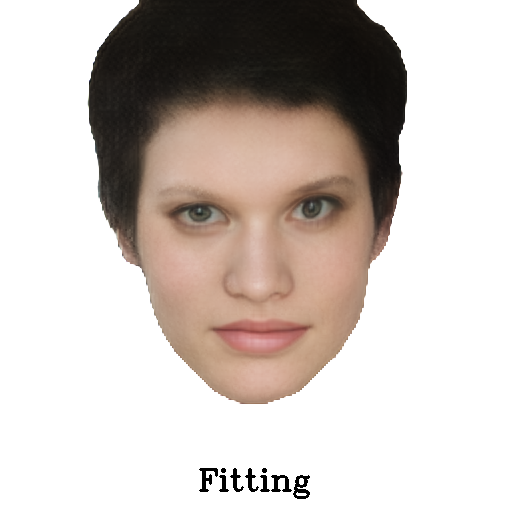}
\end{minipage}

\begin{minipage}[t]{0.212\linewidth}
\centering
\includegraphics[width=\linewidth]{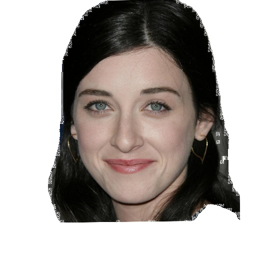}
\end{minipage}
\begin{minipage}[t]{0.212\linewidth}
\centering
\includegraphics[width=\linewidth]{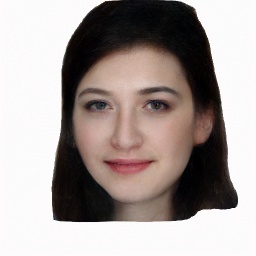}
\end{minipage}
\begin{minipage}[t]{0.212\linewidth}
\centering
\includegraphics[width=\linewidth]{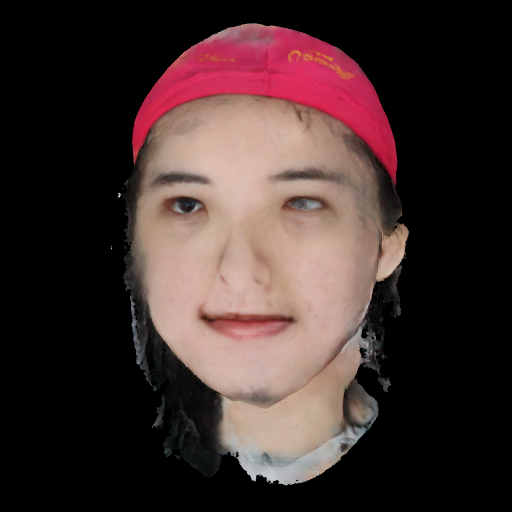}
\end{minipage}
\begin{minipage}[t]{0.212\linewidth}
\centering
\includegraphics[width=\linewidth]{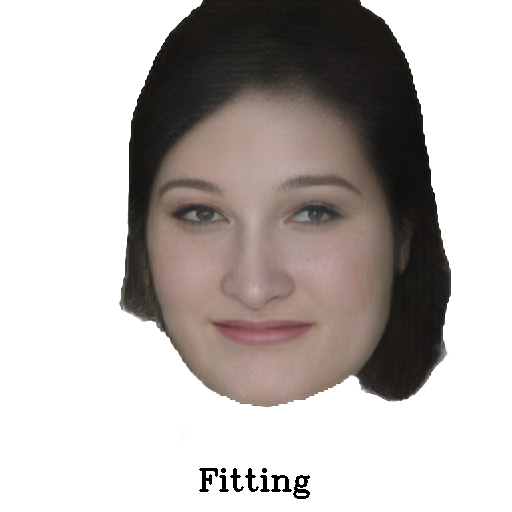}
\end{minipage}

\begin{minipage}[t]{0.212\linewidth}
\centering
\includegraphics[width=\linewidth]{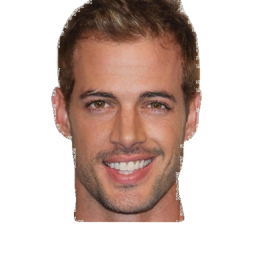}
\end{minipage}
\begin{minipage}[t]{0.212\linewidth}
\centering
\includegraphics[width=\linewidth]{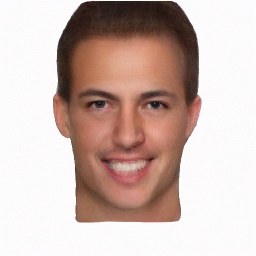}
\end{minipage}
\begin{minipage}[t]{0.212\linewidth}
\centering
\includegraphics[width=\linewidth]{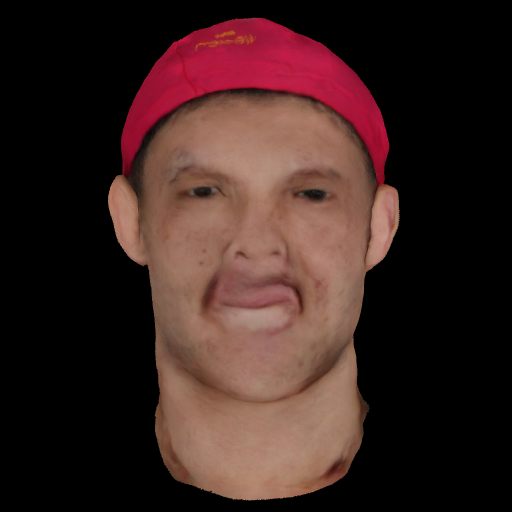}
\end{minipage}
\begin{minipage}[t]{0.212\linewidth}
\centering
\includegraphics[width=\linewidth]{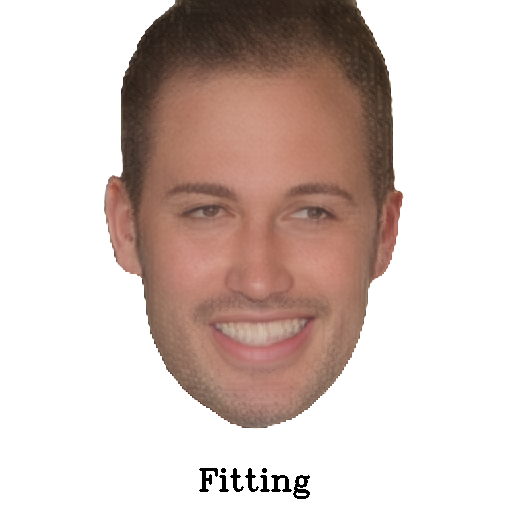}
\end{minipage}


\begin{minipage}[t]{0.212\linewidth}
\centering
\includegraphics[width=\linewidth]{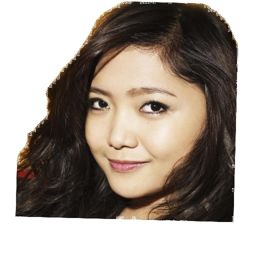}
\end{minipage}
\begin{minipage}[t]{0.212\linewidth}
\centering
\includegraphics[width=\linewidth]{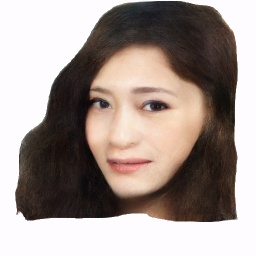}
\end{minipage}
\begin{minipage}[t]{0.212\linewidth}
\centering
\includegraphics[width=\linewidth]{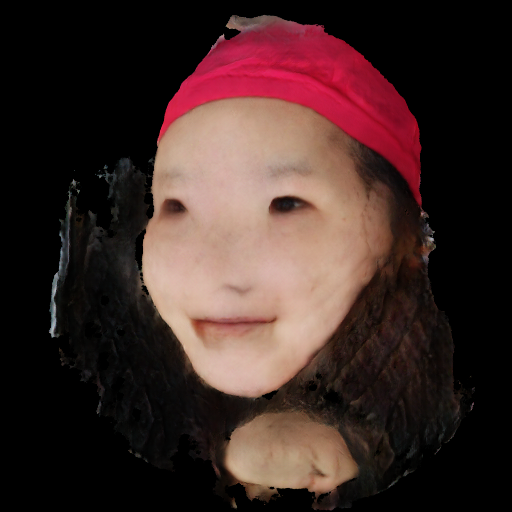}
\end{minipage}
\begin{minipage}[t]{0.212\linewidth}
\centering
\includegraphics[width=\linewidth]{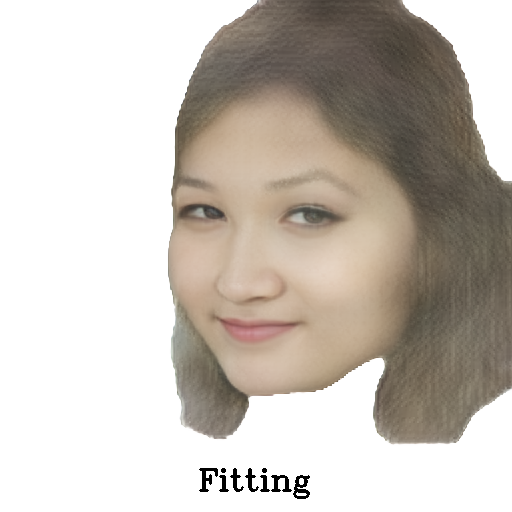}
\end{minipage}

\begin{minipage}[t]{0.212\linewidth}
\centering
\includegraphics[width=\linewidth]{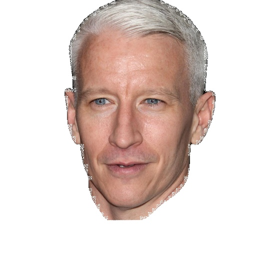} \\
\small a) Target
\end{minipage}
\begin{minipage}[t]{0.212\linewidth}
\centering
\includegraphics[width=\linewidth]{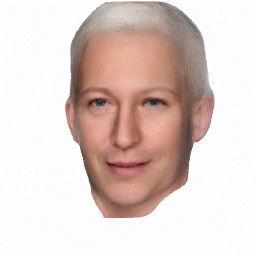} \\
\small b) Proposed
\end{minipage}
\begin{minipage}[t]{0.212\linewidth}
\centering
\includegraphics[width=\linewidth]{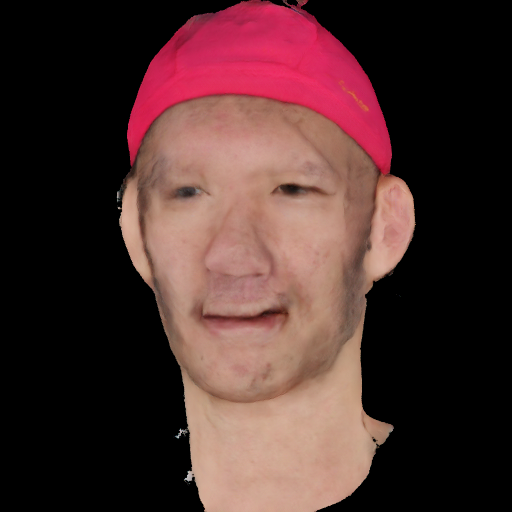} \\
\small c) MoFaNeRF~\cite{zhuang2022mofanerf}
\end{minipage}
\begin{minipage}[t]{0.212\linewidth}
\centering
\includegraphics[width=\linewidth]{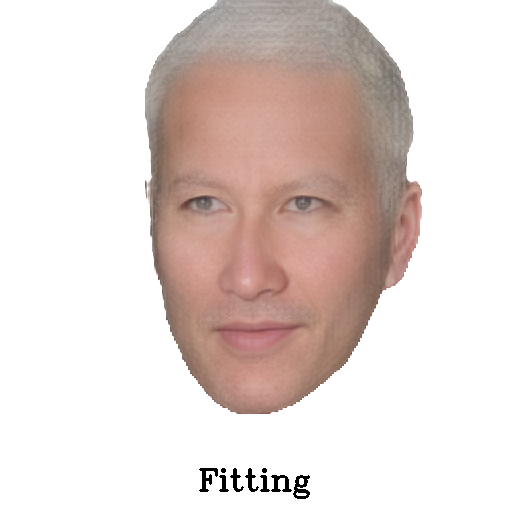} \\
\small d) HeadNeRF~\cite{hong2021headnerf}
\end{minipage}
\vspace{-0.8mm}
\caption{\small \textbf{CelebAHQ Representation Power}. We show additional results on CelebAHQ, an in-the-wild dataset that neither our method nor our baselines have trained on. Similar to FFHQ, our model demonstrates significantly higher faithfulness to the test subjects compared to the baselines.  
}\label{fig:CelebAHQRepresentationPower}
\end{center}\vspace{-4mm}
\end{figure*}

We show additional in-the-wild results for the CelebAHQ~\cite{CelebA-HQ} and FFHQ~\cite{FFHQ} datasets. As shown in Figs.~\ref{fig:FFHQRepresentationPower} and~\ref{fig:CelebAHQRepresentationPower}, INFAMOUS-NeRF achieves substantially higher representation power than the baselines on in-the-wild images. In particular, many components of the subject including hairstyle, skin tone, and expression are modeled more faithfully than the baselines. 

\begin{figure}[hbt!]
\vspace{-2mm}
\begin{center}

\begin{minipage}[t]{0.3\linewidth}
\centering
\includegraphics[width=\linewidth]{Figures/CelebAHQ_Representation_Power/18_256.png}
\end{minipage}
\begin{minipage}[t]{0.3\linewidth}
\centering
\includegraphics[width=\linewidth]{Figures/CelebAHQ_Representation_Power/18_ours.jpg}
\end{minipage}
\begin{minipage}[t]{0.3\linewidth}
\centering
\includegraphics[width=\linewidth]{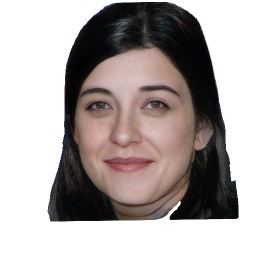}
\end{minipage}


\begin{minipage}[t]{0.3\linewidth}
\centering
\includegraphics[width=\linewidth]{Figures/CelebAHQ_Representation_Power/92_256.png}
\end{minipage}
\begin{minipage}[t]{0.3\linewidth}
\centering
\includegraphics[width=\linewidth]{Figures/CelebAHQ_Representation_Power/92_ours.jpg}
\end{minipage}
\begin{minipage}[t]{0.3\linewidth}
\centering
\includegraphics[width=\linewidth]{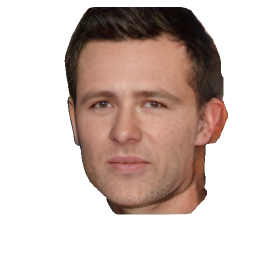}
\end{minipage}

\begin{minipage}[t]{0.3\linewidth}
\centering
\includegraphics[width=\linewidth]{Figures/CelebAHQ_Representation_Power/201_256.png} \\
\small a) Target
\end{minipage}
\begin{minipage}[t]{0.3\linewidth}
\centering
\includegraphics[width=\linewidth]{Figures/CelebAHQ_Representation_Power/201_ours.jpg} \\
\small b) Proposed
\end{minipage}
\begin{minipage}[t]{0.3\linewidth}
\centering
\includegraphics[width=\linewidth]{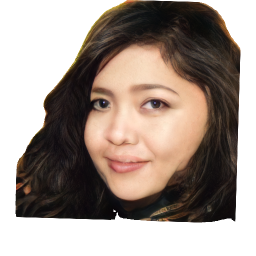} \\
\small c) E$3$DGE~\cite{lan2022e3dge}
\end{minipage}
\vspace{-0.8mm}
\caption{\small \textbf{CelebAHQ Representation Power Compared to E$3$DGE}. We show additional qualitative results on CelebAHQ comparing our representation power with E$3$DGE~\cite{lan2022e3dge}, a $3$D-aware GAN based face modeling method. Our method avoids distorting facial features, such as changing gaze direction and duplicating ears, compared to E$3$DGE. 
}\label{fig:E3DGEQualitative}
\end{center}\vspace{-4mm}
\end{figure}

\section{Ablating the Effect of the DDPM}

To further clarify that the hypernetwork is responsible for increasing representation power whereas the DDPM simply serves the role of improving the image quality, we perform two more ablations where we separately remove the hypernetwork and DDPM from INFAMOUS-NeRF. As shown in Tab.~\ref{tab:AblationsDDPM}, the improvement in representation power is coming from the hypernetwork, where all metrics improve noticeably compared to the model trained without the hypernetwork. On the other hand, when we remove the DDPM, $3$ quantitative metrics actually improve slightly whereas $1$ becomes slightly worse. This reinforces that the DDPM's role is simply to improve the perceptual quality of the rendered images. 

\section{Comparisons against $3$D-Aware GAN Methods}

\begin{table}[t!]
\caption{\small
\textbf{In-the-Wild Fitting Comparison with $3$D-Aware GANs}. Our method outperforms E$3$DGE, a $3$D-aware GAN based face modeling method, on in-the-wild fitting performance for our CelebAHQ test set. 
}\label{tab:3DAwareGANsFitting}
\vspace{-5mm}
\begin{center}
\scalebox{0.8}{
\setlength\tabcolsep{5pt}  
\begin{tabular}{c c c}
\hline
Method & PSNR & SSIM \\
\hline
E$3$DGE~\cite{lan2022e3dge} & $18.49\pm2.14$ & $0.7212\pm0.0850$ \\
\hline
INFAMOUS-NeRF (Ours) & $\mathbf{19.36\pm1.18}$ & $\mathbf{0.7710\pm0.0803}$ \\
\hline
\end{tabular}}
\end{center}
\vspace{-6mm}
\end{table} 

Although we primarily focused on comparing against baselines that have the same editing capacity as INFAMOUS-NeRF (editing appearance, shape, and expression), we also find that $3$D-aware GAN methods for face modeling are noteworthy baselines even if they are limited to only reconstruction~\cite{PanoHead,EG3D} or reconstruction and limited expression editing~\cite{lan2022e3dge} due to their focus on generalizable in-the-wild face modeling. We therefore compare with the method E$3$DGE~\cite{lan2022e3dge} on in-the-wild fitting performance using our CelebAHQ test set. We compare solely on CelebAHQ and not FFHQ since E$3$DGE trains using the entire FFHQ dataset. As shown in Tab.~\ref{tab:3DAwareGANsFitting}, our method is still able to outperform E$3$DGE~\cite{lan2022e3dge} on in-the-wild fitting. Although the gap in performance is noticeably smaller than HeadNeRF~\cite{hong2021headnerf} and MoFaNeRF~\cite{zhuang2022mofanerf} as would be expected for $3$D-aware GAN methods, we find that E$3$DGE does not always yield a completely stable solution during fitting and will sometimes distort facial features (\textit{e.g} duplicate ears, misshapen eyes, changing eye color, and changing gaze direction). See Fig.~\ref{fig:E3DGEQualitative} for qualitative comparisons. This is likely due to instability from training jointly with adversarial loss, which INFAMOUS-NeRF avoids by using a DDPM (generative model but not adversarial) rather than a GAN and separating the DDPM refinement stage from the fitting stage. 

\section{Face Geometry Evaluation}

\begin{table}[t!]
\caption{\small
\textbf{Face Geometry Evaluation on FaceScape}. We compare our learned geometry against MoFaNeRF~\cite{zhuang2022mofanerf} and the SoTA explicit face modeling method FaceVerse~\cite{wang2022faceverse}. We achieve lower RMSE than MoFaNeRF and comparable performance with FaceVerse. (NVS: novel view synthesis, Fit: 3DMM fitting). 
}\label{tab:FaceGeometryEvaluation}
\vspace{-5mm}
\begin{center}
\scalebox{0.8}{
\setlength\tabcolsep{5pt}  
\begin{tabular}{c c c}
\hline
Method & RMSE (NVS) & RMSE (Fit) \\
\hline
FaceVerse~\cite{wang2022faceverse} &  $\mathbf{0.4744\pm0.0914}$ & $\mathbf{0.4719\pm0.0835}$ \\
\hline
INFAMOUS-NeRF (Ours) & $0.5250\pm0.1010$ & $0.4922\pm0.0782$ \\
\hline
\hline
MoFaNeRF~\cite{zhuang2022mofanerf} & $0.7397\pm0.2554$ & $0.6347\pm0.1979$ \\
\hline
INFAMOUS-NeRF (Ours) & $\mathbf{0.5439\pm0.1220}$ & $\mathbf{0.4783\pm0.0873}$\\
\hline
\end{tabular}}
\end{center}
\vspace{-6mm}
\end{table} 

To evaluate the quality of our $3$D face geometry, we generate a depth map for each novel view by treating the ray point with the highest volume density dependent weight $\omega_{i}$ as the surface point for each pixel. To generate a groundtruth depth map, we project the groundtruth face shape provided by FaceScape to the target view and use z-buffering to generate the final depth map. 
We evaluate our face geometry against MoFaNeRF~\cite{zhuang2022mofanerf}, the SoTA NeRF-based face modeling method and FaceVerse~\cite{wang2022faceverse}, the SoTA mesh-based explicit face modeling method, using RMSE as the metric. As shown in Tab.~\ref{tab:FaceGeometryEvaluation}, we outperform MoFaNeRF~\cite{zhuang2022mofanerf} and achieve comparable performance as FaceVerse. 
This is expected since FaceVerse is an explicit face modeling method that can better leverage $3$D face priors such as $3$DMMs. 
We outperform MoFaNeRF in geometry estimation by making our shape latent code $\mathbf{\beta}$ larger and more expressive ($128$ dimensions) than the $50$-dim PCA coefficients offered by FaceScape, which MoFaNeRF fixes their shape latent codes to. By doing so, we allow our shape latent space to model more shape variation than $50$ constrained PCA bases. 
Furthermore, since MoFaNeRF trains jointly with adversarial loss, some of their rendered images and corresponding shapes do not converge to stable solutions. 
When evaluating, we compute the mean RMSE for the full portrait when comparing with MoFaNeRF and only the face region covered by the $3$DMM when comparing with FaceVerse to be fair to each method.

\section{Adaptive Sampling Details}
We describe our adaptive sampling strategy again in full detail, including every step of the algorithm. When sampling pixels to train NeRF methods, it is important to be conscious of the memory budget as each pixel involves casting a ray into the $3$D volume. Sampling every pixel in the image at every iteration is often infeasible given memory constraints, particularly at higher resolutions. Most NeRF methods select pixels by employing uniform sampling, which risks sacrificing fidelity due to undersampling regions with fine detail (\textit{e.g.}, the eyes, nose, and mouth) as well as oversampling nearly textureless regions such as the cheek. MoFaNeRF~\cite{zhuang2022mofanerf} proposes a landmark-based sampling scheme where $60\%$ of the pixels are sampled around the inner face landmarks and the rest are sampled uniformly. While this will naturally improve the quality of the rendered inner face, it ignores other areas where the error is generally high (\textit{e.g.}, the head boundary). In addition, uniform sampling is more redundant if some regions (\textit{e.g.}, the background) are textureless or uniform.  

In light of this, we introduce a novel adaptive sampling method that stochastically selects new pixels at each iteration based on their loss values. Our intuition is that a higher loss generally indicates that a pixel lies within a region with large gradients or fine details and should be sampled more frequently to properly maintain the fidelity. We thus want to assign certain pixels a higher probability of being sampled than others by producing an $H\!\!\times\!\!W$ probability map $\mathbf{P}_{i}$ for each image $\mathbf{I}_{i}$ that contains the per-pixel sampling probability. We describe our sampling strategy in detail in the supplementary materials. 

We thus randomly select $30$\% of the sampled pixels and for each of these pixels $(x_{i, p}, y_{i,p})$, we produce a per-pixel $2$D gaussian weight $G_{i,p}(\boldsymbol{\mu}_{i,p}, \mathbf{\Sigma}_{i,p}, \mathbf{l}_{i, c})$ defined as: 
\begin{equation}
\begin{split}
G_{i,p}(\boldsymbol{\mu}_{i,p}, \mathbf{\Sigma}_{i,p}, \mathbf{l}_{i, c}) = \frac{s_{i,p}}{2\pi |\mathbf{\Sigma}_{i,p}|^{1/2}}* \\ \exp \left\{- \frac{1}{2} \left(\mathbf{l}_{i, c}-\boldsymbol{\mu}_{i,p}\right)^{'} \mathbf{\Sigma}_{i,p}^{-1} \left(\mathbf{l}_{i, c}-\boldsymbol{\mu}_{i,p} \right)  \right\} 
\end{split}
\end{equation}
where $\mathbf{l}_{i, c}=(x_{i, c}, y_{i,c})$ is the $c^\text{th}$ pixel in $\mathbf{I}_{i}$ and $s_{i,p}$ is the value of $\mathcal{L}_\text{recon}$ at pixel $(x_{i, p}, y_{i,p})$. We can therefore accumulate a weight for every pixel $\mathbf{l}_{i,c}$ in $\mathbf{I}_{i}$ using the sum of contributions from each gaussian weight $G_{i,p}(\boldsymbol{\mu}_{i,p}, \mathbf{\Sigma}_{i,p}, \mathbf{l}_{i, c})$, where the gaussian is centered at the sampled pixel $(x_{i, p}, y_{i,p})$. Therefore, we compute the per-pixel probability map $\mathbf{P}_{i}$ for image $\mathbf{I}_{i}$ as: 
\begin{equation}
    \mathbf{P}_{i}(x_{i,c}, y_{i,c}) = \sum_{p=1}^{N}G_{i,p}(\boldsymbol{\mu}_{i,p}, \mathbf{\Sigma}_{i,p}, \mathbf{l}_{i, c}).
\end{equation}
As a final step, we normalize $\mathbf{P}_{i}$ to sum to $1$. $\mathbf{P}_{i}$ is updated every iteration and serves as the per-pixel probability distribution behind our adaptive sampling method, thus sampling more from regions with challenging textures (higher loss values) and less from near textureless regions (lower loss values). We reset $\mathbf{P}_{i}$ to the uniform distribution every $10$ iterations when image $\mathbf{I}_{i}$ is sampled to avoid converging at local regions of the image permanently and find that this yields optimal performance. 

During training, we sample $50\%$ of the rays around the inner face landmarks and the other $50\%$ using our adaptive sampling scheme. We find that this partition yields the best of both worlds: INFAMOUS-NeRF renders the inner face with high-fidelity and adaptively assigns significant weight to other regions with high reconstruction error not covered by the landmarks such as the head boundary.  
\section{Loss Functions}

Here we describe our loss functions in detail. In stage $1$, our method is supervised at the image level by a reconstruction loss $\mathcal{L}_\text{recon}$, a structural dissimilarity loss $\mathcal{L}_\text{DSSIM}$ similar to ~\cite{PhysicsGuidedRelighting},  
and a gradient loss $\mathcal{L}_\text{gradient}$ to ensure that image-level gradients are preserved in the rendered images, which we describe in detail in the supplementary materials. 

We define $\mathcal{L}_\text{recon}$ as follows: 
\begin{equation}
\begin{split}
    \mathcal{L}_\text{recon}= \frac{1}{3N}\sum_{p=1}^{N}(\|\mathbf{C}_{i,p}-\hat{\mathbf{C}}_{i,p}\|_\text{1}),
\end{split}
\end{equation}
where $N$ is the number of sampled rays per image. 
Since $\mathcal{L}_\text{DSSIM}$ and $\mathcal{L}_\text{gradient}$ inherently model local image patterns, we apply them by taking the groundtruth view and replacing any sampled pixels with our model's predicted colors, which we refer to as $\mathbf{I}^{'}_{i}$. Then $\mathcal{L}_\text{DSSIM}$ is defined as: 
\begin{equation}
    \mathcal{L}_\text{DSSIM} = \frac{1-\text{SSIM}(\mathbf{I}_{i}, \mathbf{I}'_{i})}{2},
\end{equation}
and $\mathcal{L}_\text{gradient}$ is defined as: 
\begin{equation}
    \mathcal{L}_\text{gradient} = \frac{1}{3N}(\|\nabla \mathbf{I}_{i}-\nabla \mathbf{I}'_{i}\|_\text{1}),
\end{equation}
where $\nabla$ denotes the gradient operation.

To improve the rendering quality near the face boundary, we also add our novel photometric surface rendering constraint $\mathcal{L}_\text{sur}.$ 
\begin{equation}
\begin{split}
   \mathcal{L}_\text{sur} = \frac{1}{6N}\sum_{p=1}^{N}(\|\hat{c}_{1,p,k_{1,p}}-C_{1, p}\|_\text{1}+\\\|\hat{c}_{2,p,k_{2,p}}-C_{2, p}\|_\text{1}+\|\hat{c}_{1,p,k_{1,p}}-C_{2, p_{1,2}}\|_\text{1}+\\\|\hat{c}_{2,p,k_{2,p}}-C_{1, p_{2,1}}\|_\text{1}),
\end{split}
\end{equation}
where $\hat{c}_{i,p,k_{i,p}}$ is the color predicted at the $k_{i,p}^{th}$ ray point (the predicted surface point $\mathbf{s}_{i, p}$) for the $p^{th}$ pixel sampled from the $i^{th}$ image in the batch and $k_{i,p}$ is defined as:
\begin{align}
k_{i,p}=\argmax_{j \in 1,2,....,M}\omega_{i,p,j}.
\label{eq:surfacePoint}
\end{align} 
$\mathbf{C}_{i, p}$ is the groundtruth color of the $p^{th}$ pixel in the $i^{th}$ batch image, and $\mathbf{C}_{i, p_{j,i}}$ is the groundtruth color of the pixel in the $i^{th}$ image corresponding to the surface point $\mathbf{s}_{j, p}$. 
Our final training loss for stage $1$ is therefore: 
\begin{equation}
\begin{split}
    \mathcal{L}_\text{total}=\alpha_{1}\mathcal{L}_\text{recon}+\alpha_{2}\mathcal{L}_
    \text{DSSIM}+\\\alpha_{3}\mathcal{L}_\text{gradient}+\alpha_{4}\mathcal{L}_\text{sur},
\end{split} 
\end{equation}
where $\alpha_{1}=\alpha_{2}=\alpha_{3}=\alpha_{4}=1$ are the loss weights. 
As mentioned previously, the training objective for our stage $2$ conditional DDPM is: 
\begin{align}
 \Eb{t, \bx_0, \bepsilon}{ \left\| \bepsilon - \bepsilon_\theta(\sqrt{\bar\alpha_t} \bx_0 + \sqrt{1-\bar\alpha_t}\bepsilon, t, \mathbf{I}^{'}_{i}, \mathbf{S}) \right\|^2} \label{eq:training_objective_conditional_ours_repeat}
\end{align} 
where $\mathbf{I}^{'}_{i}$ is the coarse image condition and $\mathbf{S}$ is the subject-specific noise map. 
As mentioned before, the training objective for our stage $2$ conditional DDPM is given by ~Eq.~\ref{eq:training_objective_conditional_ours}. 

\begin{table}
\begin{center}
\caption{\small
\textbf{Structure of $F_{A}$}. The network structure of our subject-specific appearance MLP (PE: positional encoding, $\mathbf{x}$: input $3$D point, $\mathbf{d}$: viewing direction, $\oplus$: concatenation). All LeakyReLU layers use a slope of $0.2$. 
}\label{tab:MLPStructuresFA}
\scalebox{0.75}{
\setlength\tabcolsep{8pt}  
\begin{tabular}{c c c c c}
\hline
Layer & Input & Input Size & Output Size & Activation \\
\hline
$A_{1}$ & $\alpha$ $\oplus$ PE($\mathbf{x}$) & $322$ & $128$ & LeakyReLU \\
\hline
$A_{2}$ & $A_{1}$ & $128$ & $128$ & LeakyReLU \\
\hline
$A_{3}$ & $A_{2}$ $\oplus$ $\alpha$ $\oplus$ PE($\mathbf{x}$) & $450$ & $128$ & LeakyReLU \\
\hline
$A_{4}$ & $A_{3}$ $\oplus$ PE($\mathbf{d}$) & $194$ & $64$ & LeakyReLU \\
\hline
$A_{5}$ & $A_{4}$ & $64$ & $3$ & Sigmoid \\
\hline
\end{tabular}}
\end{center}
\end{table}

\begin{table}
\centering
\caption{\small
\textbf{Structure of $F_{S}$}. The network structure of our subject-specific shape MLP (PE: positional encoding, $\mathbf{x}$: input $3$D point, $\oplus$: concatenation, $\mathbf{\epsilon}^{'}$=$MS$($\mathbf{\beta}$)*$\mathbf{\epsilon}$+$MB$($\mathbf{\beta}$), $MS/MB$: $3$ layer MLPs). All LeakyReLU layers use a slope of $0.2$. 
}\label{tab:MLPStructuresFS}
\scalebox{0.75}{
\setlength\tabcolsep{8pt}  
\begin{tabular}{c c c c c}
\hline
Layer & Input & Input Size & Output Size & Activation \\
\hline
$MB_{1}$ & $\mathbf{\beta}$ & $128$ & $128$ & LeakyReLU \\
\hline
$MB_{2}$ & $MB_{1}$ & $128$ & $128$ & LeakyReLU \\
\hline
$MB_{3}$ & $MB_{2}$ & $128$ & $64$ & None \\
\hline
$MS_{1}$ & $\mathbf{\beta}$ & $128$ & $128$ & LeakyReLU \\
\hline
$MS_{2}$ & $MS_{1}$ & $128$ & $128$ & LeakyReLU \\
\hline
$MS_{3}$ & $MS_{2}$ & $128$ & $64$ & None \\
\hline
$E_{1}$ & $\mathbf{\epsilon}^{'}$ $\oplus$ PE($\mathbf{x}$) & $130$ & $128$ & LeakyReLU \\
\hline
$E_{2}$ & $E_{1}$ & $128$ & $128$ & LeakyReLU \\
\hline
$E_{3}$ & $E_{2}$ & $128$ & $128$ & LeakyReLU \\
\hline
$S_{1}$ & $\mathbf{\beta}$ $\oplus$ $E_{3}$ & $256$ & $128$ & LeakyReLU \\
\hline
$S_{2}$ & $S_{1}$ & $128$ & $128$ & LeakyReLU \\
\hline
$S_{3}$ & $\mathbf{\beta}$ $\oplus$ $E_{3}$ $\oplus$ $S_{2}$ & $384$ & $128$ & LeakyReLU \\
\hline
$S_{4}$ & $S_{3}$ & $128$ & $128$ & LeakyReLU \\
\hline
$S_{5}$ & $\mathbf{\alpha}$ $\oplus$ PE($\mathbf{x}$) $\oplus$ $S_{4}$ & $450$ & $64$ & LeakyReLU \\
\hline
$S_{6}$ & $S_{5}$ & $64$ & $1$ & ReLU \\
\hline
\end{tabular}}
\end{table}

\section{Detailed Model Architecture}

We provide a more comprehensive explanation of our stage $1$ model architecture here. Our identity latent code $\theta$, our appearance latent code $\alpha$, our shape latent code $\beta$, and our expression latent code $\epsilon$ are $256$, $256$, $128$, and $64$ dimensions respectively. Next, we describe in detail the structures of $H$, $F_{A}$ and $F_{S}$ which represent our hypernetwork, our appearance MLP, and our shape MLP respectively. 

$H$ contains $2$ layers for every layer in $F_{A}$ and $F_{S}$ since it estimates subject-specific weights. The first fully connected layer is always $256\times256$ followed by LeakyReLU and takes $\theta$ as input. The second fully connected layer is $256\times P$, where $P$ is the number of parameters in a particular layer.     

We write out the architectures of $F_{A}$ and $F_{S}$ in Tabs.~\ref{tab:MLPStructuresFA} and ~\ref{tab:MLPStructuresFS}. Notice that we only require a hidden size of $128$ to achieve substantially higher representation power than MoFaNeRF, which uses a hidden size of $1024$ given the same number of training subjects. Also, we only require $5$ and $15$ layers in our appearance and shape MLPs respectively, while MoFaNeRF requires $12$ and $24$ layers. Since the memory footprint of NeRF rendering is directly correlated with the MLP sizes given each point passes through the same MLPs, our method has a substantially smaller memory footprint than MoFaNeRF (only $12$GB when rendering $1000$ rays per image during training with $128$ points per ray and a batch size of $2$). Our method can thus render significantly more rays given the same amount of memory and can be scaled up more easily to achieve better performance (\textit{e.g}, raising MLP hidden sizes, rendering more rays per iteration) without suffering as much from memory constraints.    



